\documentclass[journal]{IEEEtran}

\usepackage[utf8]{inputenc} 
\usepackage[T1]{fontenc}    
\usepackage{hyperref}       
\usepackage{url}            
\usepackage{booktabs}       
\usepackage{amsfonts}       
\usepackage{nicefrac}       
\usepackage{microtype}      
\usepackage{graphicx}
\usepackage{tabularx} 
\usepackage{mathtools, nccmath}
\usepackage{epstopdf}
\usepackage{wrapfig} 
\usepackage{caption}
\AppendGraphicsExtensions{.pdf}
\usepackage[noadjust]{cite}
\usepackage{xcolor}
\usepackage{multirow}
\usepackage{subcaption}
\usepackage{booktabs}

\newcommand{\rjc}[1]{{\textcolor[rgb]{0,0,0}{#1}}}

\begin{document}
\bstctlcite{IEEEexample:BSTcontrol}
\title{Pathomic Fusion: An Integrated Framework for Fusing Histopathology and Genomic Features for Cancer Diagnosis and Prognosis}


\author{Richard J. Chen,
        Ming Y. Lu,
        Jingwen Wang,
        Drew F. K. Williamson,
        Scott J. Rodig,\\
        Neal I. Lindeman and
        Faisal Mahmood$^*$
        
        \vspace{-3mm}
        
\noindent\thanks{{(Corresponding Author: Faisal Mahmood faisalmahmood@bwh.harvard.edu). All authors are with the Department of Pathology, Brigham and Women's Hospital, Harvard Medical School, Boston, MA. R.J.C. M.L. and F.M. are also with the Broad Institute of Harvard and MIT and the at Dana-Farber Cancer Institute, Boston, MA. R.J.C. is also with the Department of Biomedical Informatics, Harvard Medical School, Boston, MA. email: {richardchen@g.harvard.edu; \{mlu16,jwang111,dwilliamson,srodig,nlindeman\}@bwh.harvard.edu This work was supported by NIH NIGMS R35GM138216 (to F.M.), Nvidia GPU grant program, Google Cloud Research Grant. R.J.C. is funded by the National Science Foundation Graduate Fellowship and the NIH NHGRI T32HG002295 training program.
}}}

}


\maketitle

\begin{abstract}
Cancer diagnosis, prognosis, and therapeutic response predictions are based on morphological information from histology slides and molecular profiles from genomic data. However, most deep learning-based objective outcome prediction and grading paradigms are based on histology or genomics alone and do not make use of the complementary information in an intuitive manner. In this work, we propose \textit{Pathomic Fusion}, an interpretable strategy for end-to-end multimodal fusion of histology image and genomic (mutations, CNV, RNA-Seq) features for survival outcome prediction. Our approach models pairwise feature interactions across modalities by taking the Kronecker product of unimodal feature representations, and controls the expressiveness of each representation via a gating-based attention mechanism. Following supervised learning, we are able to interpret and saliently localize features across each modality, and understand how feature importance shifts when conditioning on multimodal input. We validate our approach using glioma and clear cell renal cell carcinoma datasets from the Cancer Genome Atlas (TCGA), which contains paired whole-slide image, genotype, and transcriptome data with ground truth survival and histologic grade labels. In a 15-fold cross-validation, our results demonstrate that the proposed multimodal fusion paradigm improves prognostic determinations from ground truth grading and molecular subtyping, as well as unimodal deep networks trained on histology and genomic data alone. The proposed method establishes insight and theory on how to train deep networks on multimodal biomedical data in an intuitive manner, which will be useful for other problems in medicine that seek to combine heterogeneous data streams for understanding diseases and predicting response and resistance to treatment. Code and trained models are made available at: {\color{blue} \textit{\href{https://github.com/mahmoodlab/PathomicFusion}{https://github.com/mahmoodlab/PathomicFusion}}}.
\end{abstract}



\IEEEpeerreviewmaketitle

\section{Introduction}

\IEEEPARstart{C}{ancer} diagnosis, prognosis and therapeutic response prediction is usually accomplished using heterogeneous data sources including histology slides, molecular profiles, as well as clinical data such as the patient's age and comorbidities. Histology-based subjective and qualitative analysis of the tumor microenvironment coupled with quantitative examination of genomic assays is the standard-of-care for most cancers in modern clinical settings \cite{Wen2008, Nayak2011, Aldape2015, Olar2013}. As the field of anatomic pathology migrates from glass slides to digitized whole slide images, there is a critical opportunity for development of algorithmic approaches for joint image-omic assays that make use of phenotypic and genotypic information in an integrative manner.




The tumor microenvironment is a complex milieu of cells that is not limited to only cancer cells, as it also contains immune, stromal, and healthy cells. Though histologic analysis of tissue provides important spatial and morphological information of the tumor microenvironment, the qualitative inspection by human pathologists has been shown to suffer from large inter- and intraobserver variability \cite{shanes1987interobserver}. Moreover, subjective interpretation of histology slides does not make use of the rich phenotypic information that has shown to have prognostic relevance \cite{courtiol2019deep}. Genomic analysis of tissue biopsies can provide quantitative information on genomic expression and alterations, but cannot precisely isolate tumor-induced genotypic measures and changes from those of non-tumor entities such as normal cells. \rjc{Current modern sequencing technologies such as single cell sequencing are able to resolve genomic information of individual cells in tumor specimens, with spatial transcriptomics and multiplexed immunofluoresence able to spatially resolve histology tissue and genomics together \cite{puchalski2018anatomic, jackson2020single, schapiro2017histocat, somarakis2019imacyte, abdelmoula2016data, abdelmoula2017data}. However, these technologies currently lack clinical penetration.}

\par

Oncologists often rely on both the qualitative information from histology and quantitative information from genomic data to predict clinical outcomes \cite{Gallego2015}, however, most histology analysis paradigms do not incorporate genomic information. Moreover, such methods often do not explicitly incorporate information from the spatial organization and community structure of cells, which have known diagnostic and prognostic relevance \cite{courtiol2019deep,yener2016cell,gadiya2019histographs,wang2019weakly}. Fusing morphological information from histology and molecular information from genomics provides an exciting possibility to better quantify the tumor microenvironment and harness deep learning for the development of image-omic assays for early diagnosis, prognosis, patient stratification, survival, therapeutic response and resistance prediction.

\begin{figure*}[ht]
\centering
\includegraphics[width=\textwidth]{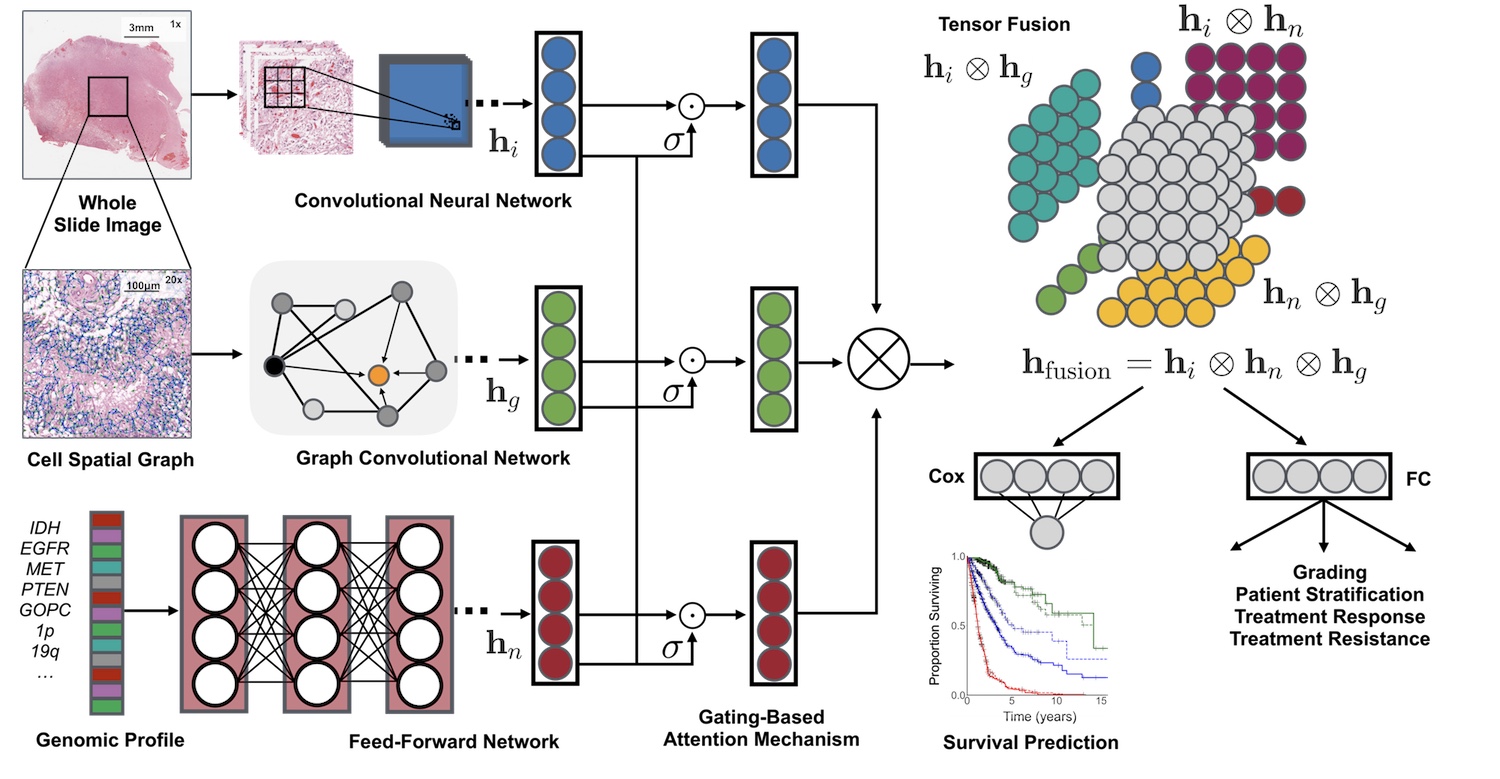}
\vspace{-6mm}
\caption{\textit{Pathomic Fusion:} An integrated framework for multimodal fusion of histology and genomic features for survival outcome prediction and classification. Histology features may be extracted using CNNs, parameter efficient GCNs or a combination of the two. Unimodal networks for the respective image and genomic features are first trained individually for the corresponding supervised learning task, and then used as feature extractors for multimodal fusion. \rjc{Multimodal fusion is performed by applying an gating-based attention mechanism to first control the expressiveness of each modality, followed by the Kronecker product to model pairwise feature interactions across modalities.}}
\vspace{-4mm}
\end{figure*}
\textbf{Contributions:} The contributions of this paper are highlighted as follows:
\begin{itemize}
    \item \textbf{Novel Multimodal Fusion Strategy:} We propose \textit{Pathomic Fusion}, a novel framework for multimodal fusion of histology and genomic features (Fig. 1). Our proposed method models pairwise feature interactions across modalities by taking the Kronecker product of gated feature representations, and controls the expressiveness of each representation using a gating-based attention mechanism.
    \item \textbf{GCNs for Cancer Outcome Prediction:} We present a novel approach for learning cell graph features in histopathology tissue using graph convolutional networks (Fig. 2), and present the first application of GCNs for cancer survival outcome prediction from histology. \rjc{GCNs act as a complementary method to CNNs for morphological feature extraction, and may be used in ileu of or in combination with CNNs during multimodal fusion for fine-grained patient stratification.}
    \item\textbf{Objective Image-Omic Quantitative Study with Multimodal Interpretability:} \rjc{In a rigorous 15-fold cross-validation-based analysis on two different disease models, we demonstrate that our image-omic fusion paradigm outperforms subjective prognostic determinations that use grading and subtyping, as well as previous state-of-the-art results for patient stratification that use deep learning. To interpret predictions made by our network in survival analysis, we use both class-activation maps and gradient-based attribution techniques to distill prognostic morphological and genomic features.}
\end{itemize}


\vspace{-2mm}
\section{Related Work}

\noindent\textbf{Survival Analysis for Cancer Outcome Prediction:} Cancer prognosis via survival outcome prediction is a standard method used for biomarker discovery, stratification of patients into distinct treatment groups, and therapeutic response prediction \cite{Zuo2019}. With the availability of high-throughput data from next-generation sequencing, statistical survival models have become one of the mainstay approaches for performing retrospective studies in patient cohorts with known cancer outcomes, with common covariates including copy number variation (CNV), mutation status, and RNA sequencing (RNA-Seq) expression \cite{cancer2011integrated, Zuo2019}. Recent work has incorporated deep learning into survival analysis, in which the covariates for a Cox model are learned using a series of fully connected layers. Yousefi \textit{et al.} \cite{Yousefi2017} proposed using stacked denoising autoencoders to learn a low dimension representation of RNA-Seq data for survival analysis, and in a follow-up work \cite{Yousefi2017}, they used Feedforward Networks to examine the relationship between gene signatures and survival outcomes. Huang \textit{et al.} \cite{huang2019salmon} proposed using weighted gene-expression network analysis as another approach for dimensionality reduction and learning eigen-features from RNA-Seq and micro-RNA data for survival analysis in TCGA. However, these approaches do not incorporate the wealth of multimodal information from heterogeneous data sources including diagnostic slides, which may capture the inherent phenotypic tumor heterogeneity that has known prognostic value.

\noindent\textbf{Multimodal Deep Learning:} Multimodal fusion via deep learning has emerged as an interdisciplinary field that seeks to correlate and combine disparate heterogeneous data modalities to solve difficult prediction tasks in areas such as visual perception, human-computer interaction, and biomedical informatics \cite{ngiam2011multimodal}. Depending on the problem, approaches for multimodal fusion range from fusion of multiview data of the same modality, such as the collection of RGB, depth and infrared measurements for visual scene understanding, to the fusion of heterogeneous data modalities, such as integrating chest X-rays, textual clinical notes, and longitudinal measurements for intensive care monitoring \cite{suresh2017clinical}. In the natural language processing community, Kim \textit{et al.} \cite{kim2016hadamard} proposed a low-rank feature fusion approach via the Hadamard product for visual question answering, often referred to as as bilinear pooling. Zadeh \textit{et al.} \cite{Zadeh2017} studied feature fusion via the Kronecker product for sentiment analysis in audio-visual speech recognition.

\noindent\textbf{Multimodal Fusion of Histology and Genomics:} Though many multimodal fusion strategies have been proposed to address the unique challenges in computer vision and natural language processing, strategies for fusing data in the biomedical domain (e.g. histology images, molecular profiles) are relatively unexplored. In cancer genomics, most works have focused on establishing correspondences between histology tissue and genomics \cite{subramanian2018correlating, idc2019ajive, coudray2018classification, kather2019deep}. For solving supervised learning tasks, previous works have generally relied on the ensembling of extracted feature embeddings from separately trained deep networks (termed late fusion) \cite{huang2019salmon, Mobadersany2018, Baltrusaitis2019}. Morbadersany \textit{et al.} \cite{Mobadersany2018} proposed a strategy for combining histology image and genomic features via vector concatenation. Cheerla \textit{et al.} \cite{Cheerla2019} developed an unsupervised multimodal encoder network for integrating histology image and genomic modalities via concatenation that is resilient to missing data. Shao \textit{et al.} \cite{shao2019integrative} proposed an ordinal multi-modal feature selection approach that identifies important features from both pathological images and multi-modal genomic data, but relies on handcrafted features from cell graph features in histology images. Beyond late fusion, there is limited work in deep learning-based multimodal learning approaches that combine histology and genomic data. \rjc{Moreover, there is little work made in interpreting histology features in these multimodal deep networks.}
 
\noindent\textbf{Graph-based Histology Analysis:} Though CNNs have achieved remarkable performance in histology image classification and feature representation, graph-based histology analysis has become a promising alternative that rivals many competitive benchmarks. The motivation for interpreting histology images as a graph of cell features (cell graph) is that these \rjc{computational morphological (morphometric)} features are more easily computed, and explicitly capture cell-to-cell interactions and their spatial organization with respect to the tissue. Prior to deep learning, previous works in learning morphological features from histology images have relied on manually constructed graphs and computing predefined statistics \cite{prewitt1979graphs}. Doyle \textit{et al.} \cite{doyle2007automated} was the first work to approach Gleason score grading in prostate cancer using Voronoi and Delauney tessellations. Shao \textit{et al.} \cite{shao2019integrative} presented an interesting approach for feature fusion of graph and molecular profile features, with graph features constructed manually and fused via vector concatenation similar to Huang \textit{et al.} \cite{huang2019salmon}. Motivated by the success of representation learning in graphs using deep networks \cite{hamilton2017inductive, lee2019self, defferrard2016convolutional,  kipf2016semi}, Anand \textit{et al.} \cite{gadiya2019histographs}, Zhou \textit{et al.}, \cite{zhou2019cgc} and Wang \textit{et al.} \cite{wang2019weakly} have used graph convolutional networks for breast, colon and prostate cancer histology classification respectively. Currently, however, there have been no deep learning based-approaches that have used graph convolutional networks for for survival outcome prediction.
\vspace{-2mm}
\section{Methods} 
Given paired histology and genomic data with known cancer outcomes, our objective is to learn a robust multimodal representation from both modalities that would outperform unimodal representations in supervised learning. Previous works have only relied on CNNs for extracting features from histology images, and late fusion for integrating image features from CNNs with genomic features. In this section, we present our novel approach for integrating histology and genomic data, \textit{Pathomic Fusion}, which fuses histology image, cell graph, and genomic features into a multimodal tensor that explicitly models bimodal and trimodal interactions from each modality. In Pathomic Fusion, histology features are extracted as two different views: image-based features using Convolutional Neural Networks (CNNs), and graph-based features using Graph Convolutional Networks (GCNs). Both networks would extract similar morphological features, however, cell graphs from histology images are a more explicit feature representation that directly model cell-to-cell interactions and cell community structure. Following the construction of unimodal features, we propose a gating-based attention mechanism that controls the expressiveness of each feature before constructing the multimodal tensor. \rjc{The objective of the multimodal tensor is to capture the space of all possible interactions between features across all modalities, with the gating-based attention mechanism used to regularize unimportant features. In subsections A-C, we describe our approach for representation learning in each modality, with subsections D-E describing our multimodal learning paradigm and approach for interpretability}. Additional implementation and training details are found in Appendix B. 






\begin{figure*}[ht]
\centering
\includegraphics[width=\textwidth]{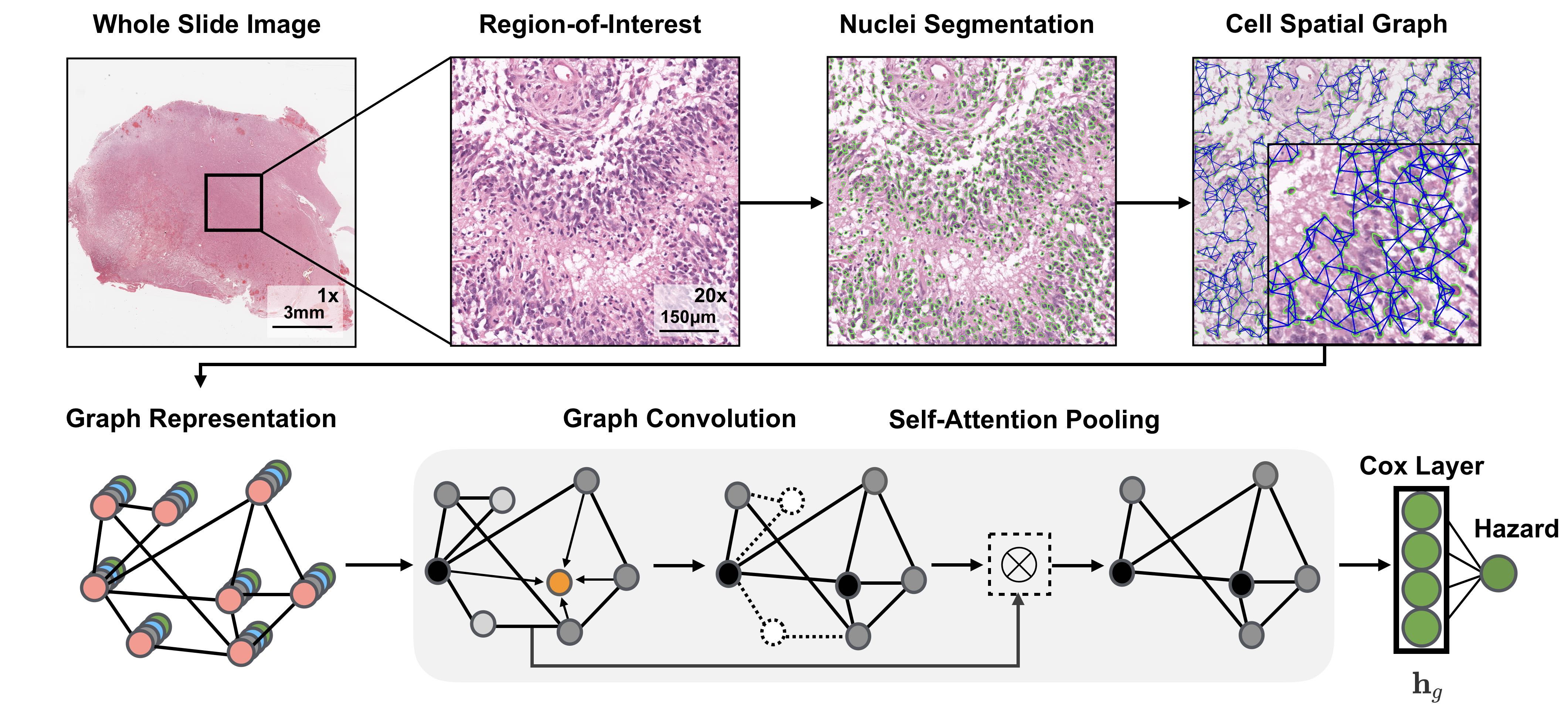}
\caption{Graph Convolutional Network for learning morphometric cell features from histology images. We represent cells in histology tissue as nodes in a graph, where cells are isolated using a deep learning-based nuclei segmentation algorithm and the connections between cells are made using KNN. Features for each cell are initialized using handcrafted features as well as deep features learned using contrastive predictive coding. The aggregate and combine functions are adopted from the GraphSAGE architecture, with the node masking and hierarchical pooling strategy adopted from SAGEPool.}
\vspace{-4mm}
\end{figure*}

\vspace{-3mm}

\subsection{Learning Patient Outcomes from H\&E Histology Tissue Images using Convolutional Neural Networks}  
Anatomic pathology has the ability to reveal the inherent phenotypic intratumoral heterogeneity of cancer, and has been an important tool in cancer prognosis for the past century \cite{Marusyk2012, yang2019guided, sari2018unsupervised, bandi2018detection}. Tumor microenvironment features such as high cellularity and microvascular proliferation have been extensively linked to tumor suppressor deficiency genes and angiogenesis, and recognized to have clinical implications in the recurrence and proliferation of cancer \cite{Guadagno2016}. To capture these features, we train a Convolutional Neural Network (CNN) on $512 \times 512$ image regions-of-interest (ROIs) at $20\times$ magnification \rjc{(0.5 $\mu$m/pixel)} as representative regions of cancer pathology. The network architecture of our CNN is VGG19 with batch normalization, which we finetuned using pre-existing weights trained on ImageNet. We extract a $\textbf{h}_i \in \mathbb{R}^{32 \times 1}$ embedding from the last hidden layer of our Histology CNN, which we use as input into Pathomic Fusion. This network is supervised by the Cox partial likelihood loss for survival outcome prediction, and cross entropy loss for grade classification (Supplement, Appendix A).

\subsection{Learning Morphometric Cell and Graph Features using Graph Convolutional Networks}

The spatial heterogeneity of cells in histopathology has potential in informing the invasion and progression of cancer and in bioinformatics tasks of interest such as cancer subtyping, biomarker discovery and survival outcome prediction \cite{Marusyk2012, Mikkelsen2018}. Unlike image-based feature representation of histology tissue using CNNs, cell graph representations explicitly capture only pre-selected features of cells, which can be scaled to cover larger regions of histology tissue.


Let $G=(V,E)$ denote a graph with nodes $V$ and edges $E$. We define $X \in \mathbb{R}^{N \times F}$ as a feature matrix of $N$ nodes in $V$ with $F$-dimensional features, and $A \in \mathbb{R}^{N \times N}$ as the adjacency matrix that holds the graph topology. To construct graphs that would capture the tumor microenvironment (Fig 2), on the same histology ROI used as input to our CNN, we 1): perform semantic segmentation to detect and spatially localize cells in a histopathology region-of-interest to define our set of nodes $V$, 2): use K-Nearest Neighbors to find connections between adjacent cells to define our set of edges $E$, 3): calculate handcrafted and deep features for each cell that would define our feature matrix $X$, and 4): use graph convolutional networks to learn a robust representation of our entire graph for survival outcome prediction.

\noindent\textbf{Nuclei Segmentation:} Accurate nuclei segmentation is important in defining abnormal cell features such as nuclear atypia, abundant tumor cellularity, and other features that would be indicative of cancer progression \cite{naylor2018segmentation, xu2015stacked, su2016robust, jia2017constrained}. Previous works rely on conventional fully convolutional networks that minimize a pixel-wise loss \cite{Kumar2017}, which can cause the network to segment multiple nuclei as one, leading to inaccurate feature extraction of nuclei shape and community structure. To overcome this issue, we use the same conditional generative adversarial network (cGAN) from our previous work to learn an appropriate loss function for semantic segmentation, which circumvents manually engineered loss functions \cite{mahmood2018adversarial, goodfellow2014generative, isola2017image}. As described in our previous work \cite{mahmood2018adversarial}, the conditional GAN framework consists of two networks (a generator $G$ and a discriminator $D$) that compete against each other in a min-max game to respectively minimize and maximize the objective $\min _{G} \max _{D} \mathcal{L}(G, D)$. Specifically, $G$ is a segmentation network that learns to translate histology tissue images $n$ into realistic segmentation masks $m$, and $D$ is a binary classification network that aims to distinguish real and predicted pairs of tissue ($(n,m)$ vs. $(n, S(n)$). Our generator is supervised with a $\mathcal{L}$1 loss and adversarial loss function, in which the adversarial loss penalizes the generator for producing segmentation masks that are unrealistic.

\begin{equation*}
\begin{aligned}
\mathcal{L}_{\mathrm{GAN}}\left(S, D_{M}\right) =& \mathbb{E}_{m, n \sim p_{\mathrm{data}}(m, n)}\left[\log D_{M}(m, n)\right] \\ 
+& \mathbb{E}_{n \sim p_{\mathrm{data}}(n)}\left[\log \left(1-D_{M}(m, S(n))\right)\right]
\end{aligned}
\end{equation*}

\noindent\textbf{Cell Graph Construction:} From our segmented nuclei, we use the $K$-Nearest Neighbors (KNN) algorithm from the Fast Library for Approximate Nearest Neighbours (FLANN) library to construct the edge set and adjacency matrix of our graph \cite{muja2009fast} (Fig 2). We hypothesize that adjacent cells will have the most significant cell-cell interactions and limit the adjacency matrix to $K$ nearest neighbours. In our investigations, we used $\mathcal{K}=5$ to detect community structure and model cellular interactions. Using KNN, our adjacency matrix $A$ is defined as:
\begin{equation*}
A_{i j}\left\{\begin{array}{cc}{1} & {\text { if } j \in \text{KNN}(i) \text { and } D(i, j)<d} \\ {0} & {\text { otherwise }}\end{array}\right.
\end{equation*}

\noindent\textbf{Manual Cell Feature Extraction:} \rjc{For each cell, we computed eight contour features (major axis length, minor axis length, angular orientation, eccentricity, roundness, area, and solidity), as well as four texture features from gray-level co-occurence matrices (GLCM) (dissimilarity, homogeneity, angular second moment, and energy). Contours were obtained from segmentation results in nuclei segmentation, and GLCMs were calculated from $64 \times 64$ image crops centered over each contour centroid. These twelve features were selected for inclusion in our feature matrix $X$, as they would describe abnormal morphological features about glioma cells such as atypia, nuclear pleomorphism, and hyperchromatism.}

\noindent\textbf{Unsupervised Cell Feature Extraction using Contrastive Predictive Coding:} \rjc{Besides manually computed statistics, we also used an unsupervised technique known as contrastive predictive coding (CPC) \cite{cpc1, cpc2, lu2019semi} to extract 1024-dimensional features from tissue regions of size 64 $\times$ 64 centered around each cell in a spatial graph.} Given a high-dimensional data sequence $\{x_t\}$ ($256 \times 256$ image crop from the histology ROI), CPC is designed to capture high-level representations shared among different portions ($64 \times 64$ image patches) of the complete signal. The encoder network $g_{enc}$ transforms each data observation $x_i$ into a low-dimensional representation $z_i$ and learns via a contrastive loss whose optimization leads to maximizing the mutual information between the available context $c_t$, computed from a known portion of the encoded sequence $\{z_i\}, i\leq t$ and future observations $z_{t+k}, k>0$. By minimizing the CPC objective, we are able to learn rich feature representations shared among various tissue regions that are specific to the cells in the underlying tissue site. Examples include the morphology and distinct arrangement of different cell types, inter-cellular interactions, and the microvascular patterns surrounding each cell. \rjc{To create CPC features for each cell, we encode $64 \times 64$ image patches centered over the centroid of each cell. These features are concatenated with our handcrafted features during cell graph construction.}

\noindent\textbf{Graph Convolutional Network:}
Similar to CNNs, GCNs learn abstracts feature representations for each feature in a node via message passing, in which nodes iteratively aggregate feature vectors from their neighborhood to compute a new feature vector at the next hidden layer in the network \cite{kipf2016semi}. The representation of an entire graph can be obtained through pooling over all the nodes, which can then be used as input for tasks such as classification or survival outcome prediction. Such convolution and pooling operations can defined as follows:

\begin{equation*}
\begin{aligned}
    a_{v}^{(k)} &= \textbf{AGGREGATE}^{(k)}\left(\left\{h_{u}^{(k-1)} : u \in \mathcal{N}(v)\right\}\right) \\ h_{v}^{(k)} &= \textbf{COMBINE}^{(k)}\left(h_{v}^{(k-1)}, a_{v}^{(k)}\right)
\end{aligned}
\end{equation*}

where $h_v^{(k)}$ is the feature vector of node $v$ at the $k-1$-th iteration of the neighborhood aggregation, $a_v^{(k)}$ is the feature vector of node $v$ at the next iteration, and AGGREGATE and COMBINE are functions for combining feature vectors between hidden layers. As defined in Hamilton \textit{et al.}, we adopt the  AGGREGATE and COMBINE definitions from GraphSAGE  \cite{hamilton2017inductive}, which for a given node, represents the next node hidden layer as the concatenation of the current hidden layer with the neighborhood features:

\begin{equation*}
\begin{aligned}
    a_{v}^{(k)} &= \textbf{MAX}\left(\left\{\textbf{ReLU}\left(W \cdot h_{u}^{(k-1)}\right), \forall u \in \mathcal{N}(v)\right\}\right) \\
    h_{v}^{(k)} &= W \cdot\left[h_{v}^{(k-1)}, a_{v}^{(k)}\right]
\end{aligned}
\end{equation*}

\rjc{Unlike other graph-structured data, cell graphs exhibit a hierarchical topology, in which the degree of eccentricity and clustered components of nodes in a graph define multiple views of how cells are organized in the tumor micro-environment: from fine-grained views such as local cell-to-cell interactions, to coarser-grained views such as structural regions of cell invasion and metastasis. In order to encode the hierarchical structure of cell graphs, we adopt the self-attention pooling strategy SAGPOOL presented in Lee \textit{et al.} \cite{lee2019self}, which is a hierarchical pooling method that performs local pooling operations of node embeddings in a graph. In attention pooling, the contribution of each node embedding in the pooling receptive field to the next network layer is adaptively learned using an attention mechanism.} The attention score $Z \in \mathbb{R}^{N \times 1}$ for nodes in $G$ can be calculated as such:

\begin{equation*}
\begin{aligned}
    Z=\sigma\left(\textbf{SAGEConv}\left(X, A+A^{2}\right)\right)
\end{aligned}
\end{equation*}

where $X$ are the node features, $A$ is the adjacency matrix, and SAGEConv is the convolution operator from GraphSAGE. To also aggregate information from multiple scales in the nuclei graph topology, we also adopt the hierarchical pooling strategy in Lee \textit{et al.} \cite{lee2019self}. Since we are constructing cell graphs on the entire image, no patch averaging of predicted hazards needs to be performed. At the last hidden layer of our Graph Convolutional SNN, we pool the node features into a $\textbf{h}_g \in \mathbb{R}^{32 \times 1}$ feature vector, which we use as an input to Pathomic Fusion.

\begin{figure*}[ht]
\centering
\includegraphics[width=\textwidth]{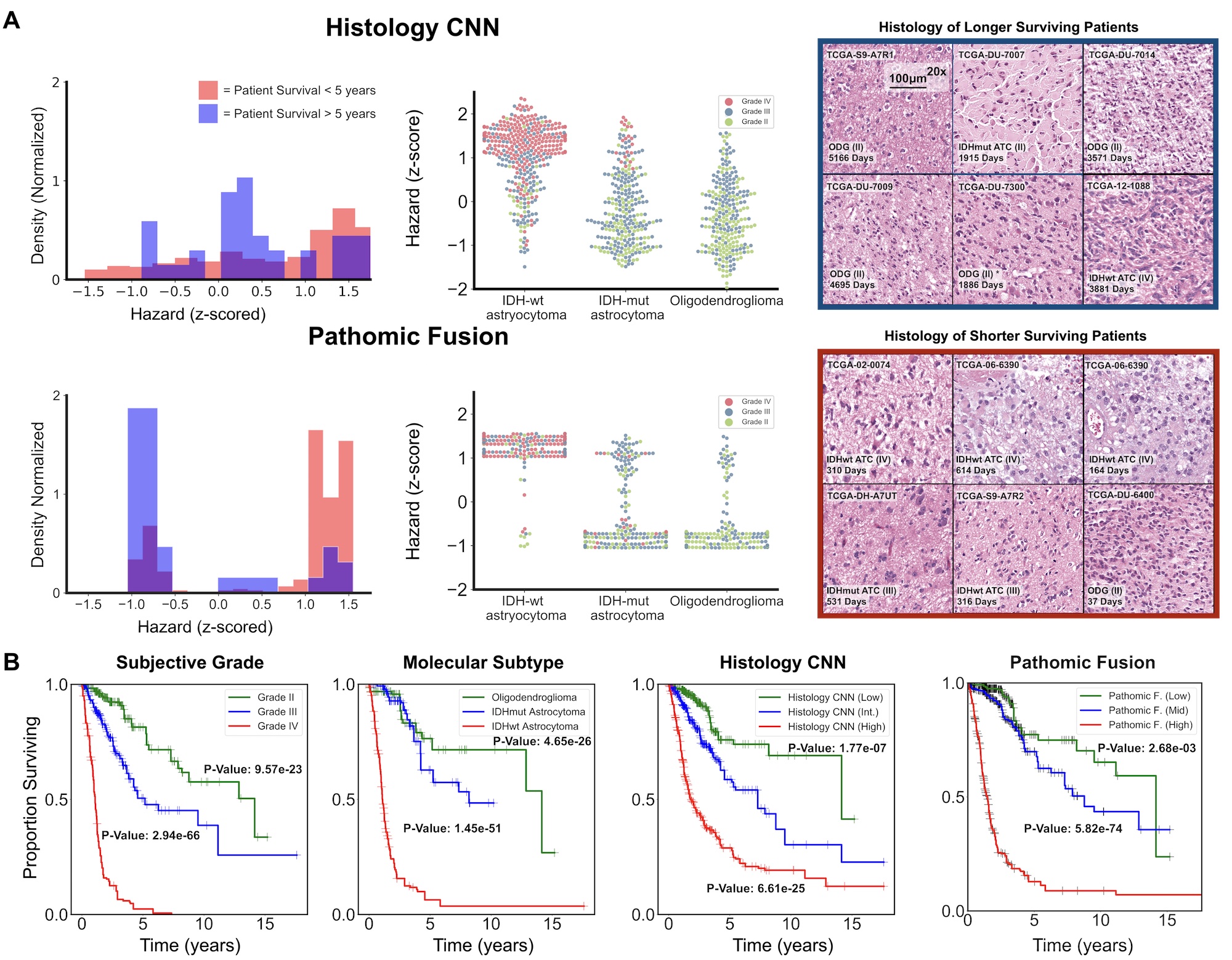}
\caption{\rjc{\textbf{Pathomic Fusion Applied to Glioblastoma and Lower Grade Glioma.} \textbf{A.} Glioma hazard distributions} amongst shorter vs. longer surviving uncensored patients and molecular subtypes \rjc{for} Histology CNN and Pathomic Fusion. Patients are defined as shorter surviving if patient death is observed before 5 years of the first follow-up (shaded red), and longer surviving if patient death is observed after 5 years of the first follow-up (shaded blue). Pathomic Fusion predicts hazard in more concentrated clusters than Histology CNN, while the distribution of hazard predictions from Histology CNN have longer tails and are more varied across molecular subtypes. In analyzing the types of glioma in the three high density regions revealed from Pathomic Fusion, we see that these regions corroborate with the WHO paradigm for stratifying patients into IDHwt ATC, IDHmut ATC, and ODG (Appendix C, Table IV). \rjc{\textbf{B.} Kaplan-Meier comparative analysis of using grade, molecular subtype, Histology CNN and Pathomic Fusion in stratifying patient outcomes. Hazard predictions from Pathomic Fusion show better stratification of mid-to-high risk patients than Histology CNN, and low-to-mid risk patients than molecular subtyping, which follows the WHO paradigm. Low / intermediate / high risk are defined by the 33-66-100 percentile of hazard predictions. Overlayed Kaplan-Meier estimates of our network predictions with WHO Grading is shown in the supplement (Appendix C, Fig. 9).}}
\vspace{-5mm}
\end{figure*}

\begin{figure*}[ht]
\centering
\includegraphics[width=\textwidth]{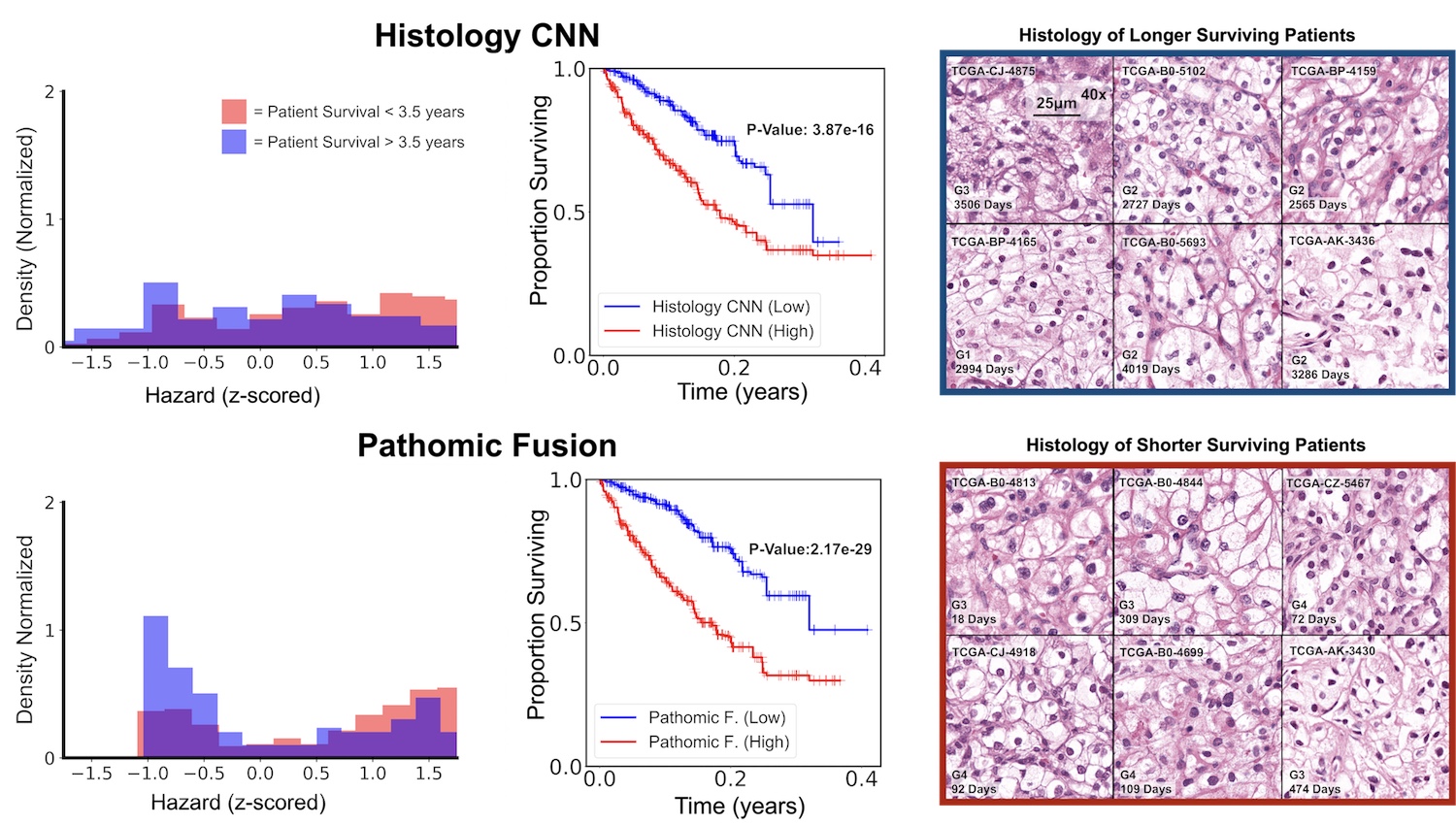}
\caption{\rjc{\textbf{Pathomic Fusion Applied to Clear Cell Renal Cell Carcinoma. CCRCC hazard distributions} amongst shorter vs. longer surviving uncensored patients for Histology CNN and Pathomic Fusion. Patients are defined as shorter surviving if patient death is observed before 3.5 years of the first follow-up (shaded red), and longer surviving if patient death is observed after 3.5 years of the first follow-up (shaded blue). Pathomic Fusion was observed to able to stratify longer and shorter surviving patients better than Histology CNN, exhibiting a bimodal distribution in hazard prediction. Overlayed Kaplan-Meier estimates of our network predictions with WHO Grading is shown in the supplement (Appendix C, Fig. 10).}}
\vspace{-4mm}
\end{figure*}

\subsection{Predicting Patient Outcomes from Molecular Profiles using Self-Normalizing Networks}
Advances in next-generation sequencing data have allowed for the profiling of transcript abundance (RNA-Seq), copy number variation (CNV), mutation status, and other molecular characterizations at the gene level, and have been frequently used to study survival outcomes in cancer. For example, isocitrate dehydrogenase 1 (IDH1) is a gene that is important for cellular metabolism, epigenetic regulation and DNA repair, with its mutation associated with prolonged patient survival in cancers such as glioma. \rjc{Other genes include EGFR, VEGF and MGMT, which are implicated in angiogenesis, which is the process of blood vessel formulation that also allows cancer to proliferate to other areas of tissue.}

\rjc{For learning scenarios that have hundreds to thousands of features with relatively few training samples, Feedforward networks are prone to overfitting. Compared to other kinds of neural network architectures such as CNNs, weights in Feedforward networks are shared and thus more sensitive training instabilities from  perturbation and regularization techniques such as stochastic gradient descent and Dropout. To mitigate overfitting on high-dimensional low sample size genomics data and employ more robust regularization techniques when training Feedforward networks, we adopt the normalization layers from Self-Normalizing Networks in Klambaeur \textit{et al.} \cite{klambauer2017self}. In Self-Normalizing Networks (SNN), rectified linear unit (ReLU) activations are replaced with scaled exponential linear units (SeLU) to drive outputs after every layer towards zero mean and unit variance. Combined with a modified regularization technique (Alpha Dropout) that maintains this self-normalizing property, we are able to train well-regularized Feedforward networks that would be otherwise prone to instabilities as a result of vanishing or explosive gradients.} Our network architecture consists of four fully-connected layers followed by Exponential Linear Unit (ELU) activation and Alpha Dropout to ensure the self-normalization property. The last fully-connected layer is used to learn a representation $\textbf{h}_n \in \mathbb{R}^{32 \times 1}$, which is used as input into our Pathomic Fusion (Fig. 1).


\subsection{Multimodal Tensor Fusion via Kronecker Product and Gating-Based Attention}
\rjc{For multimodal data in cancer pathology, there exists a data heterogeneity gap in combining histology and genomic input - histology images are spatial distributed as (R, G, B) pixels in a two-dimensional grid, whereas cell graphs are defined as a set of nodes $V$ with different sized neighborhoods and edges $V$, and genomic data is often represented as a one-dimensional vector of covariates \cite{Baltrusaitis2019}}. Our motivation for multimodal learning is that the inter-modality interactions between histology and genomic features would be able to improve patient stratification into subtypes and treatment groups. For example, in the refinement of histogenesis of glioma, though morphological characteristics alone do not correlate well with patient outcomes, their semantic importance in drawing decision boundaries is changed when conditioned on genomic biomarkers such as IDH1 mutation status and chromosomal 1p19q arm codeletion \cite{Louis2016}. 

\rjc{In this work, we aim to explicitly capture these important interactions using the Kronecker Product, which model feature interactions across unimodal feature representations, that would otherwise not be explicitly captured in feedforward layers.}  Following the construction of the three unimodal feature representations in the previous subsections, we build a multimodal representation using the Kronecker product of the histology image, cell graph, and genomic features ($\textbf{h}_i, \textbf{h}_g, \textbf{h}_{n}$). The joint multimodal tensor computed by the matrix outer product of these feature vectors would capture important unimodal, bimodal and trimodal interactions of all features of these three modalities, shown in Fig. 1 and in the equation below:


%

 

\begin{equation*}
\begin{aligned}
    \mathbf{h}_{\text{fusion}}=\left[\begin{array}{l}{\mathbf{h_i}} \\ {1}\end{array}\right] \otimes\left[\begin{array}{l}{\mathbf{h_g}} \\ {1}\end{array}\right] \otimes\left[\begin{array}{l}{\mathbf{h_n}} \\ {1}\end{array}\right]
\end{aligned}
\end{equation*}

where $\otimes$ is the outer product, and $\textbf{h}_{\text{fusion}}$ is a differential multimodal tensor that forms in a 3D Cartesian space. In this computation, every neuron in the last hidden layer in the CNN is multiplied by every other neuron in the last hidden layer of the SNN, and subsequently multiplied with every other neuron in the last hidden layer of the GCN. To preserve unimodal and bimodal feature interactions when computing the trimodal interactions, we append 1 to each unimodal feature representation. For feature vectors of size $[33 \times 1], [33 \times 1]$ and $[33 \times 1]$, the calculated multimodal tensor would have dimension $[33 \times 33 \times 33]$, where the unimodal features ($\mathbf{h_i}, \mathbf{h_g}, \mathbf{h_n}$) and bimodal feature interactions $(\mathbf{h_i} \otimes \mathbf{h_g}, \mathbf{h_g} \otimes \mathbf{h_n}, \mathbf{h_i} \otimes \mathbf{h_n})$ are defined along the outer dimension of the 3D tensor, and the trimodal interactions (captured as $\mathbf{h_i} \otimes \mathbf{h_g} \otimes \mathbf{h_n}$) in the inner dimension of the 3D tensor (Fig. 1). \rjc{Following the computation of this joint representation, we learn a final network using fully-connected layers using the multimodal tensor as input, supervised with the previously defined Cox objective for survival outcome prediction and cross-entropy loss for grade classification. Ultimately, the value of Pathomic Fusion is fusing heterogeneous modalities that have disparate structural dependencies. Our multimodal network is initialized with pretrained weights from the unimodal networks, followed by end-to-end fine-tuning of the Histology GCN and Genomic SNN.} 

\rjc{To decrease the impact of noisy unimodal features during multimodal training, before the Kronecker Product, we employed a gating-based attention mechanism that controls the expressiveness of features of each modality \cite{Arevalo2017}. In fusing histology image, cell graph, and genomic features, some of the captured features may have high collinearity, in which employing a gating mechanism can reduce the size of the feature space before computing the Kronecker Product}. For a modality $m$ with a unimodal feature representation $\textbf{h}_m$, we learn a linear transformation $W_{ign \rightarrow m}$ of modalities $\textbf{h}_{i}, \textbf{h}_{g}, \textbf{h}_{n}$ that would score the relative importance of each feature in $m$, denoted as $\mathbf{z}_m$ in the equation below.

\vspace{-2mm}

\begin{equation*}
\begin{aligned}
 \mathbf{h}_{m, \text{gated}} &= \mathbf{z}_m * \mathbf{h}_m, \forall m \in \{i,g,n\}\\ \text{where,  }
    \mathbf{h}_m &= \text{ReLU}(W_m \cdot \mathbf{h}_m)\\
    \mathbf{z}_m &= \sigma (W_{ign \rightarrow m} \cdot [\mathbf{h}_i, \mathbf{h}_g, \mathbf{h}_n])
\end{aligned}
\vspace{-2mm}
\end{equation*}

$\mathbf{z}_{m}$ can be interpreted as an attention weight vector, in which modalities $i,g,n$ attend over each feature in modality $m$. $W_m$ and $W_{ign \rightarrow m}$ are weight matrix parameters we learn for feature gating. After taking the softmax probability, we take the element-wise product of features $\textbf{h}_m$ and scores $\mathbf{z}_m$ to calculate the gated representation.

\begin{figure*}[ht]
\centering
\includegraphics[width=\textwidth]{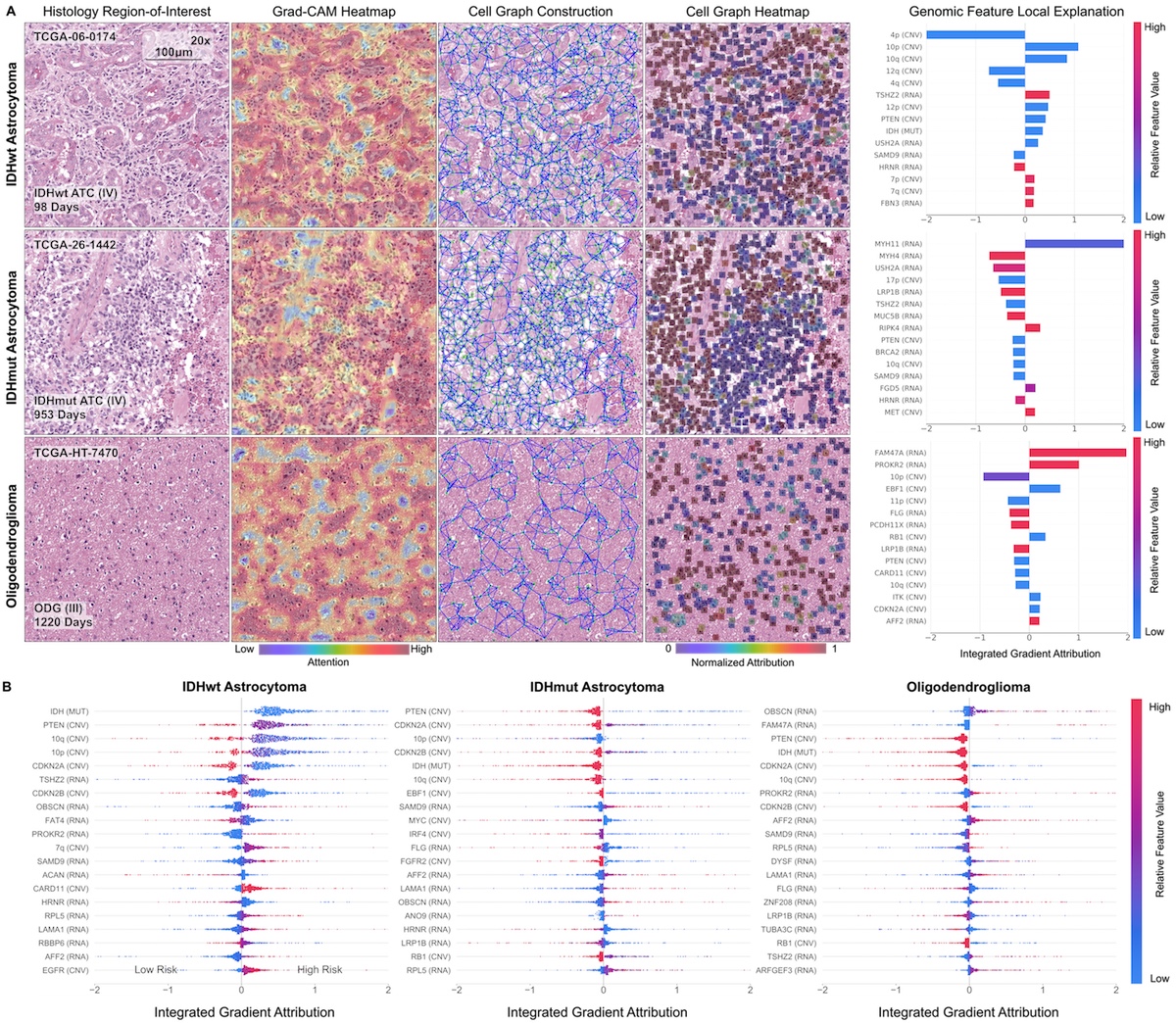}
\caption{\rjc{Multimodal interpretability by Pathomic Fusion in glioma. \textbf{A}. Local explanation of histology image, cell graph, and genomic modalities for individual patients of three molecular subtypes. In IDHwt ATC, the network detects endothelial cells of the microvascular proliferation in the histology image, while the cell graph localizes glial cells between the microvasculature. In IDHmut ATC, we observe similar localization of tumor cellularity in both the histology image and cell graph, however, attribution direction for IDH is flipped to have positive impact on survival. In ODG, we observe both modalities localizing towards different regions containing "fried egg cells" that are canonical in ODG. For each of these patients, local explanation reveals the most important genomic features used for prediction. \textbf{B}. Global explanation of top 20 genomic features for each molecular subtype in glioma. Canonical oncogenes in glioma such as IDH, PTEN, MYC and CDKN2A are attributed highly as being important for risk prediction.}}
\vspace{-6mm}
\end{figure*}

\vspace{-3mm}
\subsection{\rjc{Multimodal Interpretability}}

\rjc{To interpret our network, we modified both Grad-CAM and Integrated Gradients for visualizing image saliency feature importance across multiple types of input. Grad-CAM is a gradient-based localization technique used to produce visual explanations in image classification, in which neurons whose gradients have positive influence on a class of interest are used to produce a coarse heatmap \cite{selvaraju2017grad}. Since the last layer of our network is a single neuron for outputting hazard, we modified the target to perform back-propagation on the single neuron. As a result, the visual explanations from our network correspond with image regions used in predicting hazard (values ranging from [-3,3]). For Histology GCN and Genomic SNN, we used Integrated Gradients (IG), a gradient-based feature attribution method that attributes the prediction of deep networks to their inputs \cite{sundararajan2017axiomatic}. Similar to previous attribution-based methods such as Layer-wise Relevance Propagation \cite{bach2015pixel}, IG calculates the gradients of the input tensor $x$ across different scales against a baseline $x_i$ (zero-scaled), and then uses the Gauss-Legendre quadrature to approximate the integral of gradients.}


$$
\rjc{\text{IG}_{i}(x)::=\left(x_{i}-x_{i}^{\prime}\right) \times \int_{\alpha=0}^{1} \frac{\partial F\left(x^{\prime}+\alpha \times\left(x-x^{\prime}\right)\right)}{\partial x_{i}} d \alpha}
$$

\rjc{To adapt IG to graph-based structures, we treat the nodes in our graph input as the batch dimension, and scale each node in the graph by the number of integral approximation steps. With multimodal inputs, we can approximate the integral of gradients for each data modality.}

\section{Experimental Setup}
\subsection{Data Description}
\rjc{To validate our proposed multimodal paradigm for integrating histology and genomic features, we collected glioma and clear cell renal cell carcinoma data from the TCGA, a cancer data consortium that contains paired high-throughput genome analysis and diagnostic whole slide images with ground-truth survival outcome and histologic grade labels. For astrocytomas and glioblastomas in the merged TCGA-GBM and TCGA-LGG (TCGA-GBMLGG) project, we used $1024 \times 1024$ region-of-interests (ROIs) from diagnostic slides curated by \cite{Mobadersany2018}, and used sparse stain normalization \cite{Vahadane2016} to match all images to a standard H\&E histology image. Multiple region-of-interests (ROIs) from diagnostic slides were obtained for some patients, creating a total of 1505 images for 769 patients. 320 genomic features from CNV (79), mutation status (1), and bulk RNA-Seq expression from the top 240 differentially expressed genes (240) were curated from the TCGA and the cBioPortal \cite{Cerami2012} for each patient. For clear cell renal cell carcinoma in the TCGA-KIRC project we used manually extracted $512 \times 512$ ROIs from diagnostic whole slide images. For 417 patients in CCRCC, we collected 3 $512 \times 512$ 40x ROIs per patient, yielding 1251 images total that were similarly normalized with stain normalization. We paired these images with 357 genomic features from CNV of genes with alteration frequency greater than 7\% (117) and RNA-Seq from the top 240 differentially expressed genes (240). It should be noted that for TCGA-GBMLGG had approximately 40\% of the patients had missing RNA-Seq expression. Details regarding genomic features and data alignment of histology and genomics data are found in the implementation details (Appendix B). Our experimental setup is also described in the reproducibility section of our GitHub repository.}

\subsection{Quantitative Study}
\noindent\textbf{TCGA-GBMLGG:}  \rjc{Gliomas are a form of brain and spinal cord tumors defined by both hallmark histopathological and genomic heterogeneity in the tumor microenvironment, as well as response-to-treatment heterogeneity in patient outcomes.} The current World Health Organization (WHO) Paradigm for glioma classification stratifies diffuse gliomas based on morphological and molecular characteristics: glial cell type (astrocytoma, oligodendroglioma), IDH1 gene mutation status and 1p19q chromosome codeletion status \cite{Louis2016}. WHO Grading is made by the manual interpretation of histology using pathological determinants for malignancy (WHO Grades II, III, and IV). These characteristics form three categories of gliomas which have been extensively correlated with survival: 1) IDH-wildtype astrocytomas (IDHwt ATC), 2) IDH-mutant astrocytomas (IDHmut ATC), and 3) IDH-mutant and 1p/19q-codeleted oligodendrogliomas (ODG). IDHwt ATCs (predominantly WHO grades III and IV) have been shown to have the worst patient survival outcomes, while IDHmut ATCs (mixture of WHO Grades II, III, and IV) and ODGs (predominantly WHO Grades II and III) have more favorable outcomes (listed in increasing order) \cite{Louis2016}. \rjc{As a baseline against standard statistical approaches / WHO paradigm for survival outcome prediction, we trained Cox Proportion Hazard Models using age, gender, molecular subtypes and grade as covariates.}

In our experimentation, we conducted an ablation study comparing model configurations and fusion strategies in a 15-fold cross validation on two supervised learning tasks for glioma: 1) survival outcome prediction, and 2) cancer grade classification. For each task, we trained six different model configurations from the combination of available modalities in the dataset. First, we trained three different unimodal networks: 1) a CNN for in histology image input (Histology CNN), 2) a GCN for cell graph input (Histology GCN), and 3) a SNN for genomic features input (Genomic SNN). For cancer grade classification, we did not use mRNA-Seq expression due to missing data, lack of paired training examples, and because grade is solely determined from histopathologic appearance. After training the unimodal networks, we trained three different configurations of Pathomic Fusion: 1) GCN$\otimes$SNN, 2) CNN$\otimes$SNN, 3) GCN$\otimes$CNN$\otimes$SNN. To test for ensembling, we train multimodal networks that fused histology data with histology data, and genomic features with genomic features. We compare our fusion approach to internal benchmarks and the previous state-of-the-art \cite{Mobadersany2018} approach for survival outcome prediction in glioma, which concatenates histology ROIs with IDH1 and 1p19q genomic features. To compare with their results, we used their identical train-test split, which was created using a 15-fold Monte Carlo cross-validation \cite{Mobadersany2018}.

\noindent \rjc{\textbf{TCGA-KIRC:} Clear cell renal cell carcinoma (CCRCC) is the most common type of renal cell carcinoma, originating from cells in the proximal convoluted tubules. Histopathologically, CCRCC is characterized by diverse cystic grown patterns of cells with clear or eosinophilic cytoplasm, and a network of thin-walled "chicken wire" vasculature \cite{shuch2014pathologic, linehan2003genetic}. Genetically, it is characterized by a chromosome 3p arm loss and mutation status of the von Hippel-Lindau (VHL) gene, which leads to lead to stabilization of hypoxia inducible factors that lead to malignancy \cite{schraml2002vhl}. Though CCRCC is well-characterized, methods for staging CCRCC suffer from large intra-observer variability in visual histopathological examination. The Fuhrman Grading System for CCRCC is a nuclear grade that ranges from G1 (round or nuform nuclei with absent nucleoli) to G4 (irregular and multilobular nuclei with prominent nucleoli). At the time of the study, the TCGA-KIRC project used the Fuhrman Grading System to grade CCRCC in severity from G1 to G4, however, the grading system has received scrutiny in having poor overall agreement amongst pathologists on external cohorts \cite{shuch2014pathologic}. As a baseline against standard statistical approaches, we trained Cox Proportion Hazard Models using age, gender, and grade as covariates.}


\rjc{Similar to the ablation study conducted with glioma, we compared model configurations and fusion strategies in a 15-fold cross validation on CCRCC, and tested for ensembling effects. In demonstrating the effectiveness of Pathomic Fusion in stratifying CCRCC, we use the Fuhrman Grade as a comparative baseline in survival analysis, however, we do not perform ablation experiments on grade classification. Since CCRCC does not have multiple molecular subtypes, subtyping was also not performed, however, we perform analyses on CCRCC patient cohorts with different survival durations (shorter surviving and longer surviving patients).}

\noindent \rjc{\textbf{Evaluation:} We evaluate our method with standard quantitative and statistical metrics \rjc{for survival outcome prediction and grade classification}. For survival analysis, we evaluate all models using the Concordance Index (c-Index), which is defined as the fraction of all pairs of samples whose predicted survival times are correctly ordered among all uncensored samples (Table I, II). On glioma and CCRCC respectively, we separate the predicted hazards into 33-66-100 and 25-50-75-100 percentiles as digital grades, which we compared with molecular subtyping and grading. For significance testing of patient stratification, we use the Log Rank Test to measure if the difference of two survival curves is statistically significance \cite{bland2004logrank}. Kaplan-Meir estimates and predicted hazard distribution were used to visualize how models were stratifying patients. For grade classification, we evaluate our networks using Area Under the Curve (AUC), Average Precision (AP), F1-Score (micro-averaged across all classes), F1-Score (WHO Grade IV class only), and show ROC curves (Appendix C, Fig. 7). In total, we trained 480 models total in our ablation experiments using 15-fold cross validation. Implementation and training details for all networks are described in detail in Appendix A and B.}
\vspace{-3mm}
\section{Results and Discussion}

\begin{figure*}[ht]
\centering
\includegraphics[width=\textwidth]{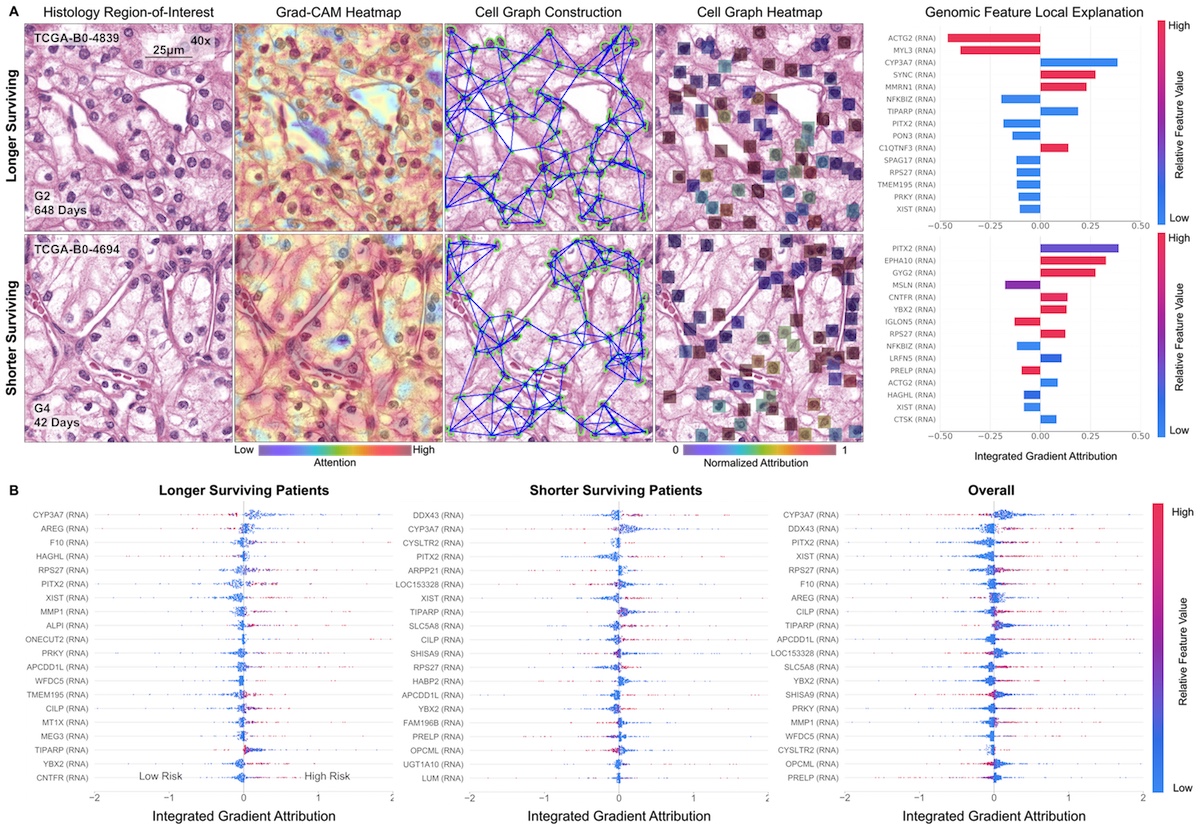}
\caption{\rjc{Multimodal interpretability by Pathomic Fusion in CCRCC. \textbf{A}. Local explanation of histology image, cell graph, and genomic modalities for two longer and shorter surviving patients. In the longer surviving patient, Pathomic Fusion localizes cells without obvious nucleoli in both the histology image and cell graph, which suggests lower-grade CCRCC and lower risk. In the shorter surviving patient, we observe Pathomic Fusion attending to large cells with prominent nucleoli and eosinophilic-to-clear cytoplasm in the cell graph, and the "chicken-wire" vasculature pattern in the histology image that is characteristic of higher-grade CCRCC. Cells without clear cytoplasms are noticeably missed in both modalities for shorter survival. For each of these patients, local explanation reveals the most important genomic features used for prediction. \textbf{B.} Global explanation of top 20 genomic features for longer surviving, shorter surviving, and all patients in CCRCC. Genes such as CYP3A7, DDX43 and PITX2 are attributed highly as being important for risk prediction, which have linked to cancer predisposition and tumor progression in CCRCC and other cancers.}}
\end{figure*}

\begin{table}
\centering
\caption{Concordance Index of Pathomic Fusion and ablation experiments in glioma survival prediction.}
\begin{tabular}{l | l}
\toprule
Model         &   c-Index\\
\midrule
Cox (Age+Gender)  & 0.732 $\pm$ 0.012\textsuperscript{*}\\
Cox (Grade)  & 0.738 $\pm$ 0.013\textsuperscript{*}\\
Cox (Molecular Subtype) & 0.760 $\pm$ 0.011\textsuperscript{*}\\
Cox (Grade+Molecular Subtype)  & 0.777 $\pm$ 0.013\textsuperscript{*}\\
\midrule
Histology CNN                &   0.792 $\pm$ 0.014\textsuperscript{*}\\
Histology GCN               &   0.746 $\pm$ 0.023\textsuperscript{*}\\
Genomic SNN               &   0.808 $\pm$ 0.014\textsuperscript{*}\\
\hline
SCNN (Histology Only) \cite{Mobadersany2018})       &  0.754\textsuperscript{*}\\
GSCNN (Histology + Genomic) \cite{Mobadersany2018})       &   0.781\textsuperscript{*}\\
\hline
\textit{Pathomic F.}  (GCN$\otimes$SNN)      &   0.812 $\pm$  0.010\textsuperscript{*}\\
\textit{Pathomic F.} (CNN$\otimes$SNN)       &   0.820 $\pm$ 0.009\textsuperscript{*}\\
\textit{Pathomic F.}  (CNN$\otimes$GCN$\otimes$SNN) & \textbf{0.826 $\pm$ 0.009}\textsuperscript{*}\\
\bottomrule
\addlinespace[1ex]
\multicolumn{1}{l}{\textsuperscript{*}$p<0.05$}
\end{tabular}
\end{table}

\begin{table}
\centering
\caption{\rjc{Concordance Index of Pathomic Fusion and ablation experiments in CCRCC survival prediction.}}
\begin{tabular}{l | c}
\toprule
Model         &   c-Index\\
\midrule
Cox (Age+Gender)  & 0.630 $\pm$ 0.024\textsuperscript{*}\\
Cox (Grade)  & 0.675 $\pm$ 0.036\textsuperscript{*}\\
\midrule
Histology CNN                &   0.671 $\pm$ 0.023\textsuperscript{*}\\
Histology GCN               &   0.646 $\pm$ 0.022\textsuperscript{*}\\
Genomic SNN               &   0.684 $\pm$ 0.025\textsuperscript{*}\\
\hline
\textit{Pathomic F.}  (GCN$\otimes$SNN)      &   0.688 $\pm$  0.029\textsuperscript{*}\\
\textit{Pathomic F.} (CNN$\otimes$SNN)       &   0.719 $\pm$ 0.031\textsuperscript{*}\\
\textit{Pathomic F.}  (CNN$\otimes$GCN$\otimes$SNN) & \textbf{0.720 $\pm$ 0.028}\textsuperscript{*}\\
\bottomrule
\addlinespace[1ex]
\multicolumn{1}{l}{\textsuperscript{*}$p<0.05$}
\end{tabular}
\vspace{-4mm}
\end{table}

\subsection{Pathomic Fusion Outperforms Unimodal Networks and the WHO Paradigm} \rjc{In combining histology image, cell graph, and genomic features via Pathomic Fusion, our approach outperforms Cox models, unimodal networks, and previous deep learning-based feature fusion approaches on image-omic-based survival outcome prediction (Table I, II). On glioma, Pathomic Fusion outperforms the WHO paradigm and the previous state-of-the-art (concatenation-based fusion \cite{Mobadersany2018}) with \textbf{6.31\%} and \textbf{5.76\%} improvements respectively, reaching a c-Index of \textbf{0.826}. In addition, we demonstrate that multimodal networks were able to consistently improve upon their unimodal baselines, with trimodal Pathomic Fusion (CNN$\otimes$GCN$\otimes$SNN) fusion of image, graph, and genomic features having the largest c-Index. Though bimodal Pathomic Fusion (CNN$\otimes$SNN) achieved similar performance metrics, the difference between low-to-intermediate digital grades ([0,33) vs. [33,66) percentile of predicted hazards) was not found to be statistically significant (Appendix C, Table III, Fig. 9). In incorporating features from GCNs, the p-value for testing difference in [0,33] vs. (33,66] percentiles decreased from 0.103 to 2.68e-03. On CCRCC, we report similar observations, with trimodal Pathomic Fusion achieving a c-Index of \textbf{0.720} and statistical significance in stratifying patients into low and high risk (Appendix C, Fig. 10). Using the c-Index metric, GCNs do not add significant improvement over CNNs alone. However, for heterogeneous cancers such as glioma, the integration of GCNs in Pathomic Fusion may provide clinical benefit in distinguishing survival curves of less aggressive tumors.}

\rjc{We also demonstrate that these improvements are not due to network ensembling, as inputting same modality twice into Pathomic Fusion resulted in overfitting (Appendix C, Table III). On glioma grade classification, we see similar improvement with Pathomic Fusion with increases of $2.75\%$ AUC, $4.23\%$ average precision, $4.27\%$ F1-score (micro), and $5.11\%$ (Grade IV) over Histology CNN, which is consistent with performance increases found in multimodal learning literature for conventional vision tasks (Appendix C, Fig. 8, Table V) \cite{Zadeh2017}.}


\subsection{\rjc{Pathomic Fusion Improves Patient Stratification}}
\rjc{To further investigate the ability of Pathomic Fusion for improving objective image-omic-based patient stratification, we plot Kaplan-Meier (KM) curves of our trained networks against the WHO paradigm (which uses molecular subtyping) on glioma (Fig. 3), and against the Fuhrman Grading System on CCRCC (Fig. 4). Overall, we observe that Pathomic Fusion allows for fine-grained stratification of survival curves beyond low vs. high survival, and that these digital grades may be useful in clinical settings in defining treatment cohorts.}

\rjc{On glioma}, similar to \cite{Mobadersany2018}, we observe that \rjc{digital grading (33-66 percentile) from} Pathomic Fusion \rjc{is} similar to that of the three defined glioma subtypes (IDHwt ATC, IDHmut ATC, ODG) that correlate with survival. In comparing Pathomic Fusion to Histology CNN, Pathomic Fusion was able to discriminate intermediate and high risk patients better than Histology CNN. Though Pathomic Fusion was \rjc{slightly worse in defining} low and intermediate risk patients, \rjc{differences between these survival curves were observed to be statistically significant (Appendix C, Table III)}. Similar confusion in discriminating low-to-intermediate risk patients is also shown in the KM estimates of molecular subtypes, which corroborates with known literature that WHO Grades II and III are more difficult to distinguish than Grades III and IV \cite{Louis2016} (Fig. 3). \rjc{In} analyzing \rjc{the} distribution of predicted hazard scores for patients in low vs. high surviving cohorts, we \rjc{also} observe that Pathomic Fusion is able to \rjc{correctly assign risk to these patients in three high-density peaks / clusters, whereas Histology CNN alone labels a majority of intermediate-to-high risk gliomas with low hazard values. In inspecting the clusters elucidated by Pathomic Fusion \rjc{([-1.0, -0.5], [1.0, 1.25] and [1.25, 1.5])}, we see that the gliomas these clusters strongly corroborate with the WHO Paradigm for stratifying gliomas into IDHwt ATC, IDHmut ATC, and ODG.}

\rjc{On CCRCC, we observe that Pathomic Fusion is able to not only differentiate between lower and higher surviving patients, but also assign digital grades that follow patient stratification by the Fuhrman Grading System (Fig. 4). Unlike Histology CNN, Pathomic Fusion is able to disentangle the survival curves of G1-G3 CCRCCs, which have overall low-to-intermediate survival. In analyzing the distribution of hazard predictions by Histology CNN, we see that risk is almost uniformly predicted across shorter and longer survival patients, which suggests that histology alone is a poor prognostic indicator for survival in CCRCC.}

\subsection{Multimodal Interpretability of Pathomic Fusion}
\rjc{In addition to improved patient stratification, we demonstrate that our image-omic paradigm is highly interpretable, in which we can attribute how pixel regions inhistology images, cells in cell graphs, and features in genomic inputs are used in survival outcome prediction.}

\rjc{ In examining IG attributions for genomic input, we were able to corroborate important markers such as IDH wildtype status in glioma and CYP3A7 under-expression in CCRCC correlating with increased risk. In glioma, our approach highlights several signature oncogenes such as PTEN, MYC, CDKN2A, EGFR and FGFR2, which are implicated in controlling cell cycle and angiogenesis (Fig. 5) \cite{2015}. In examining how feature attribution shifts when conditioning on morphological features, several genes become more pronounced in predicting survival across each subtype such as ANO9 and RB1 (Appendix C, Fig. 11). ANO9 encodes for a protein that mediates diverse physiological functions such as ion transport and phospholipid movement across the membrane. Over-expression of ANO proteins were found to be correlated with poor prognosis in many tumors, which we similarly observe in our IDHmut ATC subtype with decreased ANO9 expression decreases risk \cite{jun2017ano9}. In addition, we also observe RB1 over-expression decreases risk also in IDHmut ATC, which corroborates with known literature that RB1 is a tumor suppressor gene. Interestingly, EGFR amplication decreased in importance in IDHwt ATC, which may support evidence that EGFR is not a strong therapeutic target in glioblastoma treatment \cite{Westphal2017}. In CCRCC, Pathomic Fusion discovers decreased CYP3A7 expression and increased PITX2, DDX43, and XIST expression to correlate with risk, which have been linked to cancer predisposition and tumor progression across many cancers (Fig. 6) \cite{murray1999cytochrome, abdel2014hage, allegra2012circulating, jordan2020molecular, lee2019oncogenic}. In conditioning on morphological features, HAGHL, MMP1 and ARRP21 gene expression becomes more highly attributed in risk prediction (Appendix C, Fig 11) \cite{lin2019ghrelin, wang2019mmp}. For cancers such as CCRCC that do not have multiple molecular subtypes, Pathomic Fusion has the potential to refine gene signatures in cancers, and uncover new prognostic biomarkers that can be targeted in therapeutic treatments.} 


\rjc{Across all histology images and cell graphs in both organ types, we observe that Pathomic Fusion broadly localizes both vasculature and cell atypia as an important feature in survival outcome prediction. In ATCs, Pathomic Fusion is able to localize not only regions of tumor cellularity and microvascular proliferation in the histology image, but also glial cells between the microvasculature as depicted in the cell graph (Fig. 5). In ODG, both modalities attend towards "fried egg cells", which are mildly enlarged round cells with dark nuclei and clear cytoplasm characteristic in ODG. In CCRCC, Pathomic Fusion attends towards cells with indiscernible nucleoli in longer surviving patients, and large cells with clear nucleoli in shorter surviving patients that is indicative of CCRCC malignancy (Fig. 6). An important aspect about our method is that we are able to leverage heatmaps from both histology images and cell graphs to explain prognostic histological features used for prediction. Though visual explanations from the image and cell graph heatmap often overlap in localizing cell atypia, the cell graph can be used to uncover salient regions that are not recognized in the histology image for risk prediction. Moreover, cell graphs may have additional clinical potential in explainability, as the attributions refer to specific atypical cells rather than pixel regions.}


\section{Conclusion}


\rjc{The recent advancements made in imaging and sequencing technologies is transforming our understanding of molecular biology and medicine with multimodal data. Next-generation sequencing technologies such as RNA-Seq is redefining clinical grading paradigms to include bulk quantitative measurements from molecular subtyping \cite{Louis2016}. Tangential to this growth field has been the emergence of tissue imaging instrumentation such as whole-slide imaging, which capture the organization of cells and their surrounding tissue architecture. In this work, we present Pathomic Fusion, a novel framework for integrating data from these technologies for building objective image-omic assays for cancer diagnosis and prognosis. We extract morphological features from histology images using CNNs and GCNs and genomic features using SNNs and fuse these deep features using the Kronecker Product and a gating-based attention mechanism. We validate our approach on glioma and CCRCC data from TCGA, and demonstrate how multimodal networks in medicine can be used for fine-grained patient stratification and interpretted for finding prognostic features. The method presented is scalable and interpretable for multiple modalities of different data types, and may be used for integrating any combination of imaging and multi-omic data. The paradigm is general and may be used for predicting response and resistance to treatment. Multimodal interpretability has the ability to identify new and novel integrative bio-markers of diagnostic, prognostic and therapeutic relevance.}




{
\bibliographystyle{IEEEtran}
\bibliography{references.bib}
}
\newpage
\onecolumn

\section*{Appendix A. Deep Learning-based Survival Analysis}
Survival analysis is a task that models the time to an event, where the outcome of the event is not always observed. Such events are called censored, in which the date of the last known encounter is used as a lower bound of the survival time. For the task of cancer survival outcome prediction, an uncensored event would be patient death, and a censored event would include either patient survival or last known follow-up.


Let $T$ be a continuous random variable that represents patient survival time, and the survival function $S(t) = P(T \geq t_0)$ be the probability of a patient surviving longer than time $t_0$. We can denote the probability that an event occurs instantaneously at a time $t$ (after $t_0$) as the hazard function $\lambda(t)$. Integrating the hazard function over the time between $t$ and $t_0$ gives us the survival function \cite{cox1972regression}.

\begin{equation*}
\begin{aligned}
    \lambda(t) &= \lim _{\partial t \rightarrow 0} \frac{P(t \leq T \leq t + \partial t | T \geq t)}{\partial t}, S(t) = \exp \left(-\int_{0}^{t} \lambda(x) \partial x\right).
\end{aligned}
\end{equation*}

The most common semi-parametric approach for estimating the hazard function is the Cox proportion hazards model, which assumes that the hazard function can be parameterized as an exponential linear function $\lambda(t|x) = \lambda_0(t) e^{\beta x}$, where $\lambda_0(t)$ is the baseline hazard that describes how the risk of an event changes over time, $\beta$ are model parameters that describe how the hazard varies with covariates / features $X$ of a patient. In the original model, the baseline hazard $\lambda_0(t)$ is left unspecified, making it difficult to estimate $\beta$, however, the Cox partial log-likelihood can be derived that expresses the likelihood of an event to be observed at time $t$ for $\beta, X$ \cite{wong1986theory}.

\begin{equation*}
\begin{aligned}
    l(\beta, X) = -\sum_{i \in U}\left(X_{i} \beta-\log \sum_{j \subset R_{i}} e^{X_{j} \beta}\right), \frac{\partial l(\beta, X)}{\partial X_{i}}=\delta(i) \beta-\sum_{i, j \in C_{j}, U} \frac{\beta e^{X_{i} \beta}}{\sum_{k \in C_{j}} e^{X_{k} \beta}}
\end{aligned}
\end{equation*}

where $U$ is the set of uncensored patients, $R_i$ is the set of patients whose time of death or last follow-up is later than $i$. From the partial log-likelihood, $\beta$ can be estimated using iterative optimization algorithms such as Newton-Raphson or Stochastic Gradient Descent. To train deep networks for survival analysis, features from the hidden layer are used as covariates in the Cox model, with the derivative of the partial log-likelihood used as error during back-propagation. \rjc{To evaluate the performance of networks for survival analysis, we use the Concordance Index (c-Index), which measures the concordance of ranking of predicted hazard scores with the ground truth survival times of patients. To demonstrate how well Pathomic Fusion performs over other models, we used the c-Index as a comparative performance metric to measure how well each model is able to predict hazard scores amongst patients (higher is better). Our baseline for clinical practice was using the ground truth molecular subtypes as covariates in a Cox proportional hazard model - the canonical regression technique for modeling survival distributions. P-Values were calculated using the Log Rank Test, which we used to assess low vs. high risk stratification on all datasets, low vs. intermediate and intermediate vs. high (33-66-100 percentile) risk stratification in glioma, and 25-50-75-100 percentile risk stratification in CCRCC \cite{bland2004logrank}.}

\section*{Appendix B. Implementation Details}
\subsection{Inclusion Criteria for Genomic and Transcriptomic Features}

\begin{wrapfigure}{r}{8cm} 
    \centering
    \includegraphics[scale=0.10]{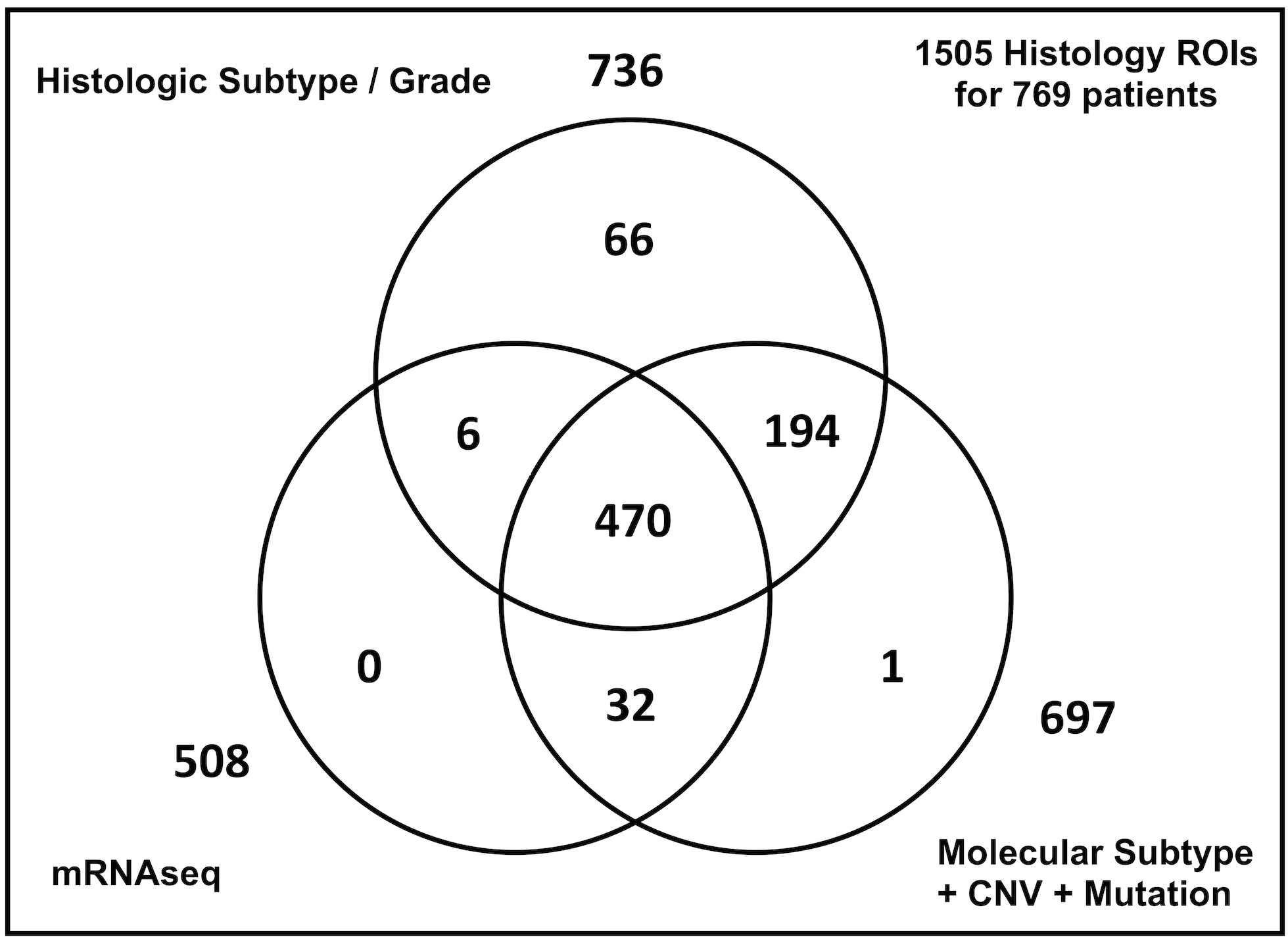}
    \caption{Missing data in the TCGA-GBMLGG project. Because not all genomic features are available for all patients, Pathomic Fusion was trained with different subsets of the data than the unimodal networks for both survival outcome prediction and grade classification.} 
    \vspace{-3mm}
\end{wrapfigure}

\indent \rjc{In our analysis on the merged TCGA-GBMLGG and TCGA-KIRC projects, we use 320 and 357 genomic features respectively. Genomic features include mutation (e.g. - binary indication of mutation status for IDH1 gene, 0/1) and copy number variation (CNV) (e.g. - amplified / deleted copies for genes and chromosomal regions). Copy number variation measurement in TCGA uses the Affymetrix SNP 6.0 array to identify repeated copies of genomic regions, with the final output as segment mean values (amplified regions have positive values, deleted regions have negative values). For TCGA-GBMLGG, mutation and CNV data used in our analysis were curated from the same set of genomic features used in Mobadersany \textit{et al.} \cite{Mobadersany2018}. Genes curated include EGFR, MDM4, MGMT, MYC and BRAF, which are implicated in oncogenic processes such as angiogenesis, apoptosis, cell growth, and differentiation. For TCGA-KIRC, we used the most amplified /deleted genes (all CNVs with greater than 7\% amplification or deletion), which yielded 117 CNV features. For both projects, we included RNA-Seq expression, which is measured as the quantified bulk abundance of mRNA transcripts. Using cBioPortal, we selected the top 240 differentially expressed genes for both projects \cite{Cerami2012}. Because our genomic features do not share any explicit spatial or temporal dependencies with each other, we feed our features through a Self-Normalizing Feedforward Network to learn a low-dimensional representation before fusing it with histology and cell graph modalities.}

\subsection{Data Missingness and Alignment in TCGA-GBMLGG}
\indent \rjc{To draw comparisons with the previous state-of-the-art, we used the existing curated TCGA-GBMLGG data found in the supplement of \cite{Mobadersany2018}, which required in careful handling of missing values in multimodal data.} For each patient, 1-3 20x $1024 \times 1024$ histology ROIs (0.5 $\mu$ / pixel) from diagnostic slides, and \rjc{320 genomic features were used.} Of the 769 patients, 72 patients have missing molecular subtype (IDH mutation and 1p19q codeletion), 33 patients have missing histological subtype and grade labels, and 256 patients have missing m\rjc{RNA-Seq} data (Fig. 6). Because multiple ROIs from diagnostic slides were obtained for some patients, each image was treated as a single data point in cross-validation, with the genomic and ground-truth label information copied over. A 15-fold Monte-carlo cross-validation was conducted using the same train-test splits as the supplement of \cite{Mobadersany2018}, which were generated randomly with 80\% training and 20\% testing (split by TCGA ID). \rjc{Depending on the task (survival prediction vs. grade classification) and combination of modalities used (histology vs. genomic vs. histology+genomic), different subsets of the train split were used to train the unimodal and multimodal networks due to missingness. In validating our models on the test splits of the cross-validation with missing data, test splits were standardized to exclude all missing data across all models (center overlap in Figure 5.). Data missingness was not an issue in working with CCRCC, with all unimodal and multimodal networks trained with the same train-test splits in a 15-fold cross-validation.}\\

\vspace{-5mm}
\subsection{Network Architectures}
\indent Three different network architectures were used to process the three modalities in our problem: 1) a VGG19 CNN with batch normalization for histology images, 2) a GCN for cell spatial graphs, and 3) a Feedforward Self-Normalizing Network for molecular profiles. The VGG19 network consists of 16 convolutional, 3 fully connected and 5 max pooling layers, with 512 $\times$ 512 sized images used as input. Dropout probabilities of 0.25 were applied after the first two fully connected layers (of size 1024), with a mild dropout (p=0.05) applied after the last hidden layer (of size 32). Our GCN consisted of 3 GraphSAGE and Self-Attention Pooling layers with hidden dimension 128, followed by two linear layers of size 128 and 32. Lastly, our Genomic SNN consists of 4 consecutive blocks of fully-connected layers with dimensions [64, 48, 32, 32], ELU activation, and Alpha Dropout. For survival outcome prediction, all networks were activated using the Sigmoid function, with the output scaled to be between -3 and 3. For grade classification, all networks were activated using the Log Softmax to compute scores for each of the 3 WHO Grades.

Our multimodal network architectures consist of two components: 1) Gating-based Modality Attention, and 2) fusion by Kronecker Product. Each modality was gated using three linear layers, with the second linear layer used to compute the attention scores. For survival outcome prediction, the genomic modality was used to gate over the image and graph modalities, while for grade classification, the histology image modality was used to gate over the genomic and graph modality. Additional dimension reduction of the gated unimodal feature representations was performed to reduce the output size of the Kronecker product feature space in the trimodal network. In our trimodal network, for survival outcome prediction, the first and third linear layers for the genomic modality have 32 hidden units to maintain the feature map dimension, with the linear layers in the image and graph modalities having 16 hidden units in order to transform the feature representations into a lower dimension. For grade classification, we maintained the feature dimension of our histology image modality instead, and reduced the dimension of the graph and genomic modalities. No feature dimension reduction was done in bimodal networks in any tasks. For feature fusion, the Kronecker product of the respective unimodal feature representations for each modality was computed, creating feature maps of size: $[33 \times 33], [33 \times 33], [33 \times 17 \times 17]$ for our CNN$\otimes$SNN, GCN$\otimes$SNN, and CNN$\otimes$GCN$\otimes$SNN. To use the unperturbed unimodal features, we appended 1 to each feature vector before computing the Kronecker Product. Dropout layers with probability $(p = 0.25)$ were inserted after gating and computing the multimodal tensor.

\subsection{Experimental Details}
\indent \rjc{Pathomic Fusion was built with PyTorch 1.5.0, PyTorch Geometric 1.5.0, Captum 0.2.0, and Lifelines 0.24.6. Node features for cell graph construction were calculated by: 1) segmenting each nuclei, 2) using the Contours Features Toolbox in in OpenCV 4.2.0, 3) the Texture Feature Toolbox in , 4) self-supervised deep features using Contrastive Predictive Coding, and 5) PyFlann 1.6.14 for graph construction.} Resources used in our experimentation include 12 Nvidia GeForce RTX 2080 Tis on local workstations, and 2 Nvidia Tesla V100s on Google Cloud.  The Histology CNN was initialized using pretrained weights from ImageNet, followed by finetuning the network using a low learning rate of 0.0005 and a batch size of 8. Random crops of 512 $\times$ 512, color jittering, and random vertical and horizontal flips were performed of data augmentation. The Histology GCN and Genomic SNN were initialized using the self-normalizing weights from Klambeur \textit{et al.} \cite{klambauer2017self}, and trained with a learning rate of 0.002 with a batch size of 32 and 64 respectively. For the Genomic SNN, a mild $\mathcal{L}_1$ regularization was also used with hyperparameter value 3e-4 to enforce feature sparsity. \rjc{All networks were trained with the same epochs using the Adam optimizer, dropout probability $p=0.25$, and a linearly decaying learning rate scheduler.}

After training the Histology CNN, for each $1024 \times 1024$ histology ROI, we extract $[32 \times 1]$ embeddings from 9 overlapping $512 \times 512$ patches, which we pair with their respective cell graph and genomic feature input as input into Pathomic Fusion. For the Histology GCN and Genomic SNN, we first trained their respective unimodal networks with the aforementioned training details, and then trained the last linear layers of the multimodal network with the unimodal network modules frozen with a learning rate of $0.0001$ and Adam solver. At epoch 5, we unfroze the genomic and graph networks, and then trained the network for 25 more epochs using a learning rate of $0.0001$, Adam solver, and a linearly decaying learning rate scheduler.\\
\vspace{-4mm}
\subsection{Evaluation Details}
\indent The predicted hazard and grade scores from each unimodal and multimodal network were evaluated on the test splits of the 15-fold cross-validation. To use the entire $1024 \times 1024$ histology image for CNN-based survival outcome prediction on TCGA-GBMLGG, similar to previous work, we computed the mean of hazard predictions from 9 overlapping $512 \times 512$ image crops across all histology ROIs belonging to each patient. For plotting the Kaplan-Meier curves, we pooled predicted hazards from all of the test splits in the 15-fold cross-validation and plotted them against their survival time. For creating the Swarm plots, we z-scored predicted hazards in each split before pooling so that scores for low vs. intermediate risk would have similar ranges in visualization. For grade classification \rjc{on TCGA-GBMLGG, we used the max softmax activation score from overlapping $512 \times 512$ patches to determine class. For CNN-based survival outcome prediction on CCRCC, we similar computed the mean of hazard predictions from $512 \times 512$ histology ROIs for each patient.}

\section*{Appendix C. Ablation Studies and Comparative Analysis}

\begin{table*}[ht]
\centering
\caption{Concordance Index \& statistical significance of Pathomic Fusion and ablation experiments in glioma survival prediction.}
\begin{tabular}{l || l l l l}
\toprule
Model         &   c-Index  $\uparrow$  & [0,50] vs. (50,100] $\downarrow$ &  [0,33] vs. (33,66] $\downarrow$ & [33,66] vs. (66,100] $\downarrow$\\
\midrule
Cox (Age+Gender)  & 0.732 $\pm$ 0.012\textsuperscript{*} & 1.90e-92 & 1.48e-38 & 5.93e-27\\
Cox (Grade)  & 0.738 $\pm$ 0.013\textsuperscript{*} & 6.00e-255 & 9.57e-23	 & 2.94e-66\\
Cox (Molecular Subtype) & 0.760 $\pm$ 0.011\textsuperscript{*} & 2.07e-228	 & 4.65e-26  & 1.45e-51\\
Cox (Grade+Molecular Subtype)       &  0.777 $\pm$ 0.013\textsuperscript{*} &            5.29e-215 &               1.14e-40 &               5.02e-52 \\
\midrule
Histology CNN                &   0.792 $\pm$ 0.014\textsuperscript{*} & 5.09e-40 & 1.77e-07 & 6.61e-25\\
Histology GCN               &   0.746 $\pm$ 0.022\textsuperscript{*} & 1.62e-21 & 2.29e-03 & 4.20e-15\\
Genomic SNN               &   0.808 $\pm$ 0.014 & 1.52e-52 & 0.153 & 2.16e-81\\
\hline
SCNN (Histology Only) \cite{Mobadersany2018})       &  0.754\textsuperscript{*} &    2.08e-61 & - & -\\
GSCNN (Histology+Genomic) \cite{Mobadersany2018})       &   0.781\textsuperscript{*} &   3.08e-64 & - & -\\
\hline
\textit{Pathomic F.}  (GCN$\otimes$SNN)      &   0.812 $\pm$  0.010\textsuperscript{*} & 1.06e-55 & 0.209 & 3.09e-80\\
\textit{Pathomic F.} (CNN$\otimes$SNN)       &   0.820 $\pm$ 0.009\textsuperscript{*} & 5.18e-57 & 0.103 & 4.56e-79\\
\textit{Pathomic F.}  (CNN$\otimes$GCN$\otimes$SNN) & \textbf{0.826 $\pm$ 0.009}\textsuperscript{*} & 7.09e-57 & 2.68e-03 & 5.82e-74\\
\bottomrule
\addlinespace[1ex]
\multicolumn{3}{l}{\textsuperscript{**}$p<0.05$}
\end{tabular}
\end{table*}

\begin{table*}[ht]
\caption{Concordance Index \& statistical significance of Pathomic Fusion and ablation experiments in CCRCC survival prediction.}
\begin{tabular}{l || l l l l l l}
\toprule
Model         &   c-Index  $\uparrow$  & [0,50] vs. (50,100] $\downarrow$ &  [0,25] vs. (25,50] $\downarrow$ & [25,50] vs. (50,75] $\downarrow$ &  [50,75] vs. (75,100] $\downarrow$\\
\midrule
Cox (Age+Gender)  & 0.630 $\pm$ 0.024\textsuperscript{*} & 1.27e-16 & 0.108 & 1.2e-05 & 0.360 \\
Cox (Grade)  & 0.675 $\pm$ 0.036\textsuperscript{*} & 4.42e-17 & 1.25e-07 & 4.52e-04 & 0.513 \\
\midrule
Histology CNN                &   0.671 $\pm$ 0.023\textsuperscript{*} & 3.87e-16 &	0.481 &	1.74e-04 &	4.19e-04\\
Histology GCN               &   0.648 $\pm$ 0.031\textsuperscript{*} & 4.23e-02 &	0.012 &	0.651 &	0.144\\
Genomic SNN               &   0.685 $\pm$ 0.024\textsuperscript{*} & 8.84e-19 & 0.480 & 0.013 &	7.07e-16\\
\hline
\textit{Pathomic F.}  (GCN$\otimes$SNN)      &   0.688 $\pm$  0.029\textsuperscript{*} & 1.65e-17 &	0.301 &	0.069 &	1.79e-12\\
\textit{Pathomic F.} (CNN$\otimes$SNN)       &   0.719 $\pm$ 0.031\textsuperscript{*} & 1.11e-27 &	0.772 &	7.00e-6 &	5.77e-12\\
\textit{Pathomic F.}  (CNN$\otimes$GCN$\otimes$SNN) & \textbf{0.720 $\pm$ 0.028}\textsuperscript{*} & 2.48e-24 &	0.087 &	9.37e-3 &	1.08e-14\\
\bottomrule
\addlinespace[1ex]
\multicolumn{3}{l}{\textsuperscript{**}$p<0.05$}
\end{tabular}
\end{table*}

\begin{figure*}[ht]
\centering
\includegraphics[width=\textwidth]{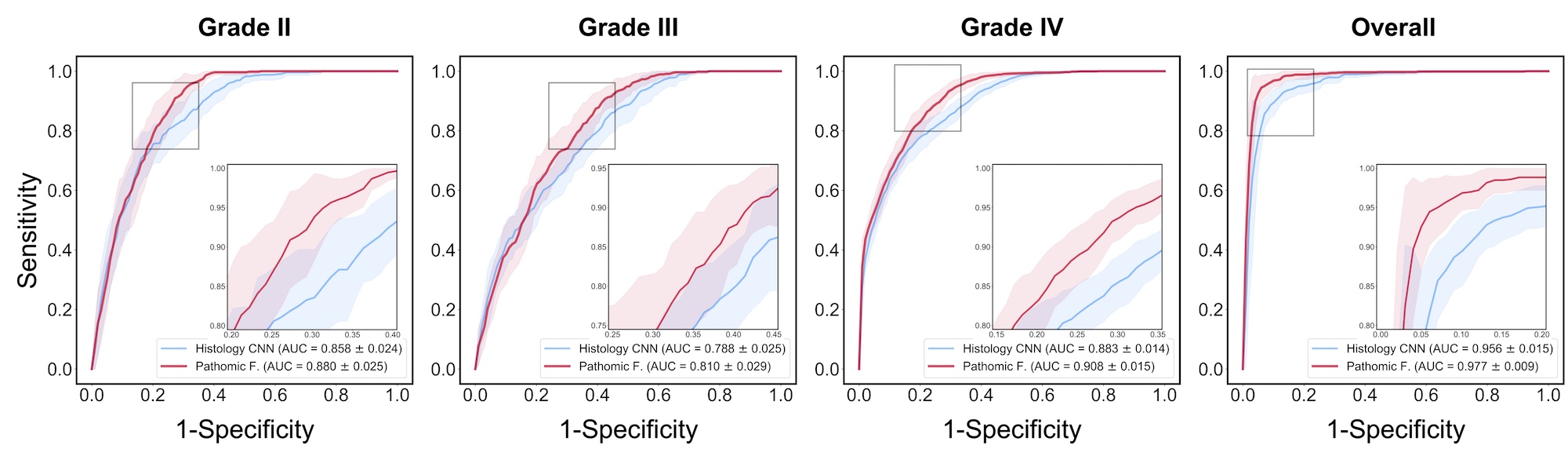}
\caption{Comparative analysis of AUC curves for Histology CNN and Pathomic Fusion in grade classification. The confidence interval is representative of the 15-fold cross validation. Since grade is usually determined via histology, our Genomic SNN that uses only CNV, gene mutation and chromosome deletion is not predictive of grade. Pathomic Fusion has greater AUCs in all cases, and performs particularly well on grade IV potentially because IDH mutation and 1p19q co-deletion from the genomic profile aid in discriminating Grade IV astrocytomas and Grade II/III oligodendrogliomas.}
\vspace{-3mm}
\end{figure*}

\subsection{Ensembling Effects}
In order to further validate the performance improvement in multimodal networks presented in Table I we conduct ensambling experiments. In other words, for a fair comparison we compare the performance of our multimodal networks against CNN$\otimes$CNN, GCN$\otimes$GCN and SNN$\otimes$SNN in order to rule out ensembling effects causing the observed improvement. Table II summarizes these results and we demonstrate that the improvement from Pathomic Fusion is greater than fusing the same modality using the same architecture. 

\begin{table*}[ht]
\caption{Comparative analysis of the ensembling effects of unimodal networks. Overall, ensemble models of Histology CNN, Histology GCN, and Genomic SNN caused networks to overfit and decrease in performance. Improvements made by ensembling were marginal compared to improvements made by Pathomic Fusion.}
\begin{tabular}{l||c c c c c | c}
\toprule
Model         &   c-Index  $\uparrow$ & AUC $\uparrow$ & AP $\uparrow$ & F1-Score (Micro) $\uparrow$ & F1-Score (Grade IV) & c-Index  $\uparrow$ \\
\midrule
Histology CNN & 0.750 $\pm$ 0.010& 0.883 $\pm$ 0.008 &  0.793 $\pm$ 0.017 &  0.717 $\pm$ 0.017 &  0.873 $\pm$ 0.013 & 0.671 $\pm$ 0.023\\
Histology (CNN + CNN) &   0.749 $\pm$ 0.010 & 0.888 $\pm$ 0.007 &  0.807 $\pm$ 0.013 &  0.715 $\pm$ 0.021 &  0.879 $\pm$ 0.015 & 0.671 $\pm$ 0.023 \\
Histology GCN & 0.722 $\pm$ 0.014 & 0.849 $\pm$ 0.011 &  0.764 $\pm$ 0.012 &  0.665 $\pm$ 0.019 &  0.849 $\pm$ 0.011 & 0.648 $\pm$ 0.031\\
Histology (GCN + GCN) & 0.720 $\pm$ 0.014 & 0.851 $\pm$ 0.015 &  0.763 $\pm$ 0.021 &  0.650 $\pm$ 0.023 &  0.812 $\pm$ 0.022 & 0.643 $\pm$ 0.027\\
Genomic SNN & 0.808 $\pm$ 0.014 & 0.853 $\pm$ 0.012 &  0.729 $\pm$ 0.018 &  0.652 $\pm$ 0.015 &  0.857 $\pm$ 0.017 & 0.684 $\pm$ 0.025\\
Genomic (SNN + SNN) & 0.794 $\pm$ 0.014 & 0.850 $\pm$ 0.012 &  0.725 $\pm$ 0.019 &  0.651 $\pm$ 0.018 &  0.856 $\pm$ 0.017 & 0.684 $\pm$ 0.025\\
\hline
\textit{Pathomic F.}  (GCN$\otimes$SNN) & 0.812 $\pm$ 0.010 &  0.897 $\pm$ 0.010 &  0.812 $\pm$ 0.016 &  0.714 $\pm$ 0.018 &  0.902 $\pm$ 0.014 & 0.686 $\pm$  0.024\\

\textit{Pathomic F.} (CNN$\otimes$SNN) & 0.820 $\pm$ 0.009 &  0.905 $\pm$ 0.010 &  \textbf{0.833 $\pm$ 0.016} &  0.730 $\pm$ 0.019 &  0.913 $\pm$ 0.011 & 0.719 $\pm$ 0.031 \\
\textit{Pathomic F.}  (CNN$\otimes$GCN$\otimes$SNN) & \textbf{0.826 $\pm$ 0.009}  & \textbf{0.908 $\pm$ 0.008} &  0.828 $\pm$ 0.016 &  \textbf{0.749 $\pm$ 0.020} &  \textbf{0.920 $\pm$ 0.014} & \textbf{0.720 $\pm$ 0.028}\\
\bottomrule
\end{tabular}
\end{table*}

\newpage


\begin{figure*}[ht]
\includegraphics[width=\textwidth]{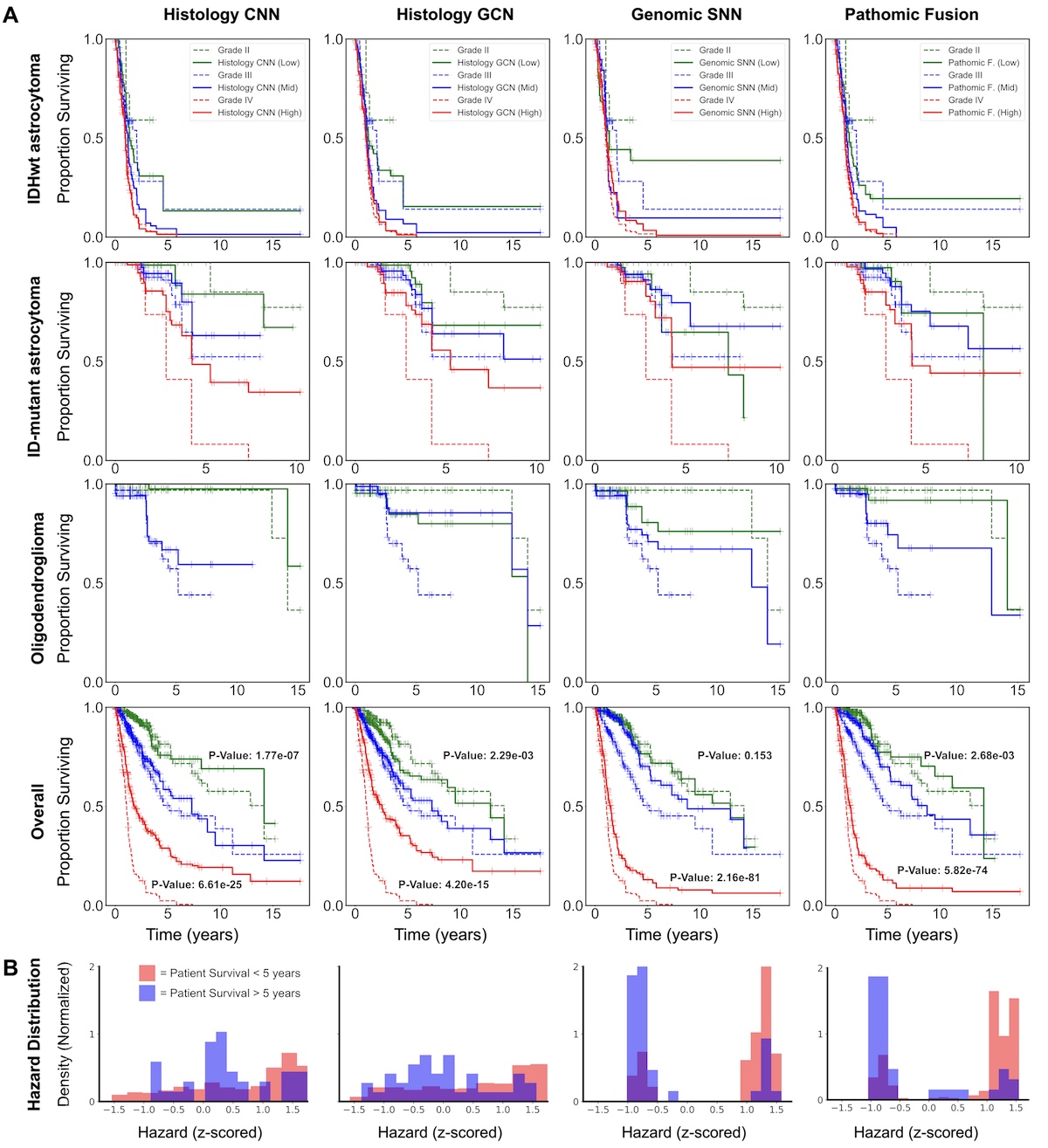}
\caption{\rjc{\textbf{A.}} TCGA-GBMLGG Kaplan-Meier comparative analysis of Histology CNN, Histology GCN, Genomic SNN, and Pathomic Fusion with respect to IDHwt ATCs, IDHmut ATCs, ODGs, and all molecular subtypes in stratifying WHO Grades II, III, and IV using the 33-66-100 percentile of hazard predictions. Overall, we observe that this heuristic has similar stratification of patients as the WHO grading system, \rjc{with Pathomic Fusion having the closest resemblance. \textbf{B.} Distribution of hazard predictions for Histology CNN, Histology GCN, Genomic SNN, and Pathomic Fusion. Histology CNN and Histology had similar skewed distributions of hazard. Qualitatively, the distribution of hazard predictions by Genomic SNN is divided into clusters, while Pathomic Fusion produced is able to delineate three clusters.}}
\end{figure*}

\begin{figure*}[ht]
\includegraphics[width=\textwidth]{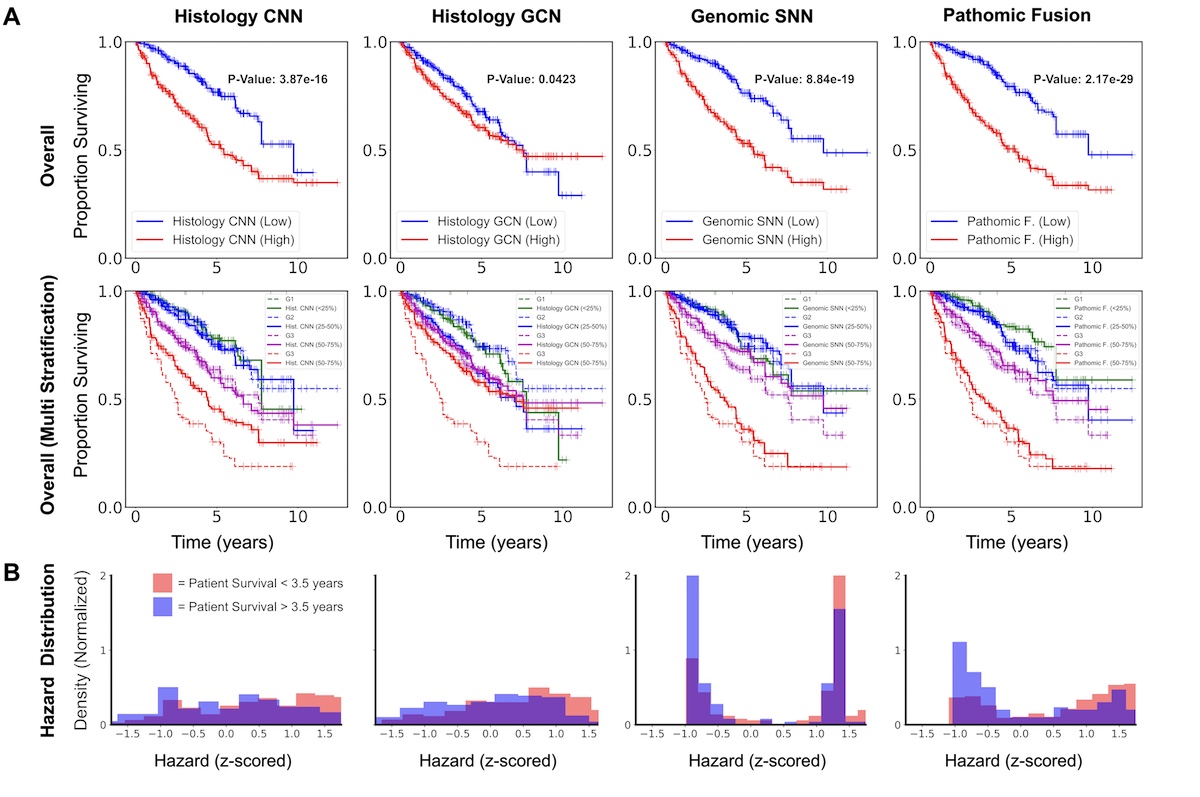}
\caption{\rjc{\textbf{A.} CCRCC Kaplan-Meier comparative analysis of Histology CNN, Histology GCN, Genomic SNN, and Pathomic Fusion in stratifying low vs. high survival using the 50-100 percentile of hazard predictions, and Fuhrman Grades I, II, III, and IV using the 25-50-75-100 percentile of hazard predictions. In performing fine-grained stratification with four risk categories, Pathomic Fusion performed the best in disentangling each category. \textbf{B.} Distribution of hazard predictions for Histology CNN, Histology GCN, Genomic SNN, and Pathomic Fusion.}}
\end{figure*}

\begin{figure*}[ht]
\includegraphics[width=\textwidth]{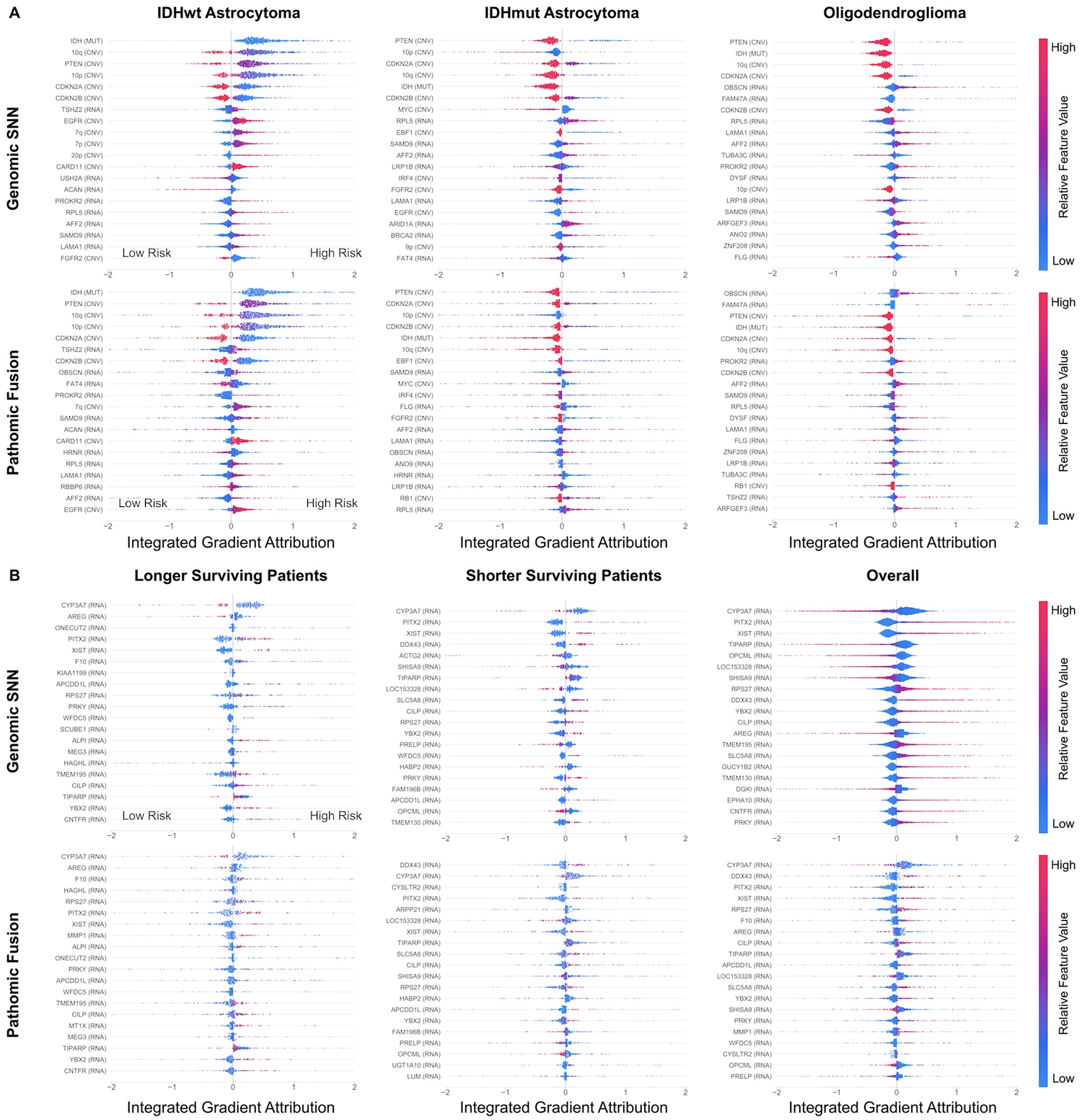}
\caption{\rjc{Genomic SNN and Pathomic Fusion global explanation across patient cohorts in TCGA-GBMLGG and CCRCC. Attribution color corresponds to low (blue) vs. high (red) feature value, and attribution direction corresponds to how the gene feature value contributes to low risk (left) vs. high risk (right). Data points in the summary plots correspond to local explanations made by Integrated Gradients when attributing features for a given sample. Top 20 features were ranked by mean absolute attribution. \textbf{A.} Global explanation for each molecular subtype in TCGA-GBMLGG. OBSCN, FAT4, HRNR, RPL5, RBBP6, RB1, and ANO9, FAM47A feature importance increased when conditioned on morphological features, while EGFR feature importance decreased. \textbf{B.} Global explanation for longer surviving, shorter surviving, and all patients in TCGA-KIRC. Longer and shorter surviving patient cohorts were defined by the top 25 longest and shortest surviving patients respectively. MMP1, HAGHL, and ARRP21 feature importance increased when conditioned on morphological features.}}
\end{figure*}


\end{document}